\PassOptionsToPackage{table}{xcolor}

\documentclass[11pt]{article}

\usepackage[final]{acl}

\usepackage{times}
\usepackage{latexsym}

\usepackage[T1]{fontenc}

\usepackage[utf8]{inputenc}

\usepackage{microtype}

\usepackage{inconsolata}

\usepackage{graphicx}

\usepackage{./palette}
\usepackage{./researchpack}

\usepackage{booktabs}
\usepackage{enumitem}
\usepackage{siunitx}
\usepackage{ulem}
\usepackage{hyperref}
\usepackage{tcolorbox}
\usepackage{pifont}
\usepackage{multirow}
\usepackage{multicol}
\usepackage[frozencache]{minted}
\usepackage{xcolor}
\usepackage{mdframed}
\usepackage{xfrac}
\usepackage{makecell}
\usepackage{circledsteps}

\newsavebox{\mathbox}

\definecolor{promptbg}{RGB}{245, 245, 245}
\definecolor{promptborder}{RGB}{200, 200, 200}
\definecolor{prompttitlebg}{RGB}{80, 80, 80}
\definecolor{prompttitlefg}{RGB}{255, 255, 255}

\newmdenv[
  backgroundcolor=promptbg,
  linecolor=promptborder,
  linewidth=1pt,
  roundcorner=4pt,
  innerleftmargin=10pt,
  innerrightmargin=10pt,
  innertopmargin=6pt,
  innerbottommargin=10pt,
  frametitlebackgroundcolor=prompttitlebg,
  frametitlefont=\bfseries\small\color{prompttitlefg},
  frametitlerule=true,
  frametitlerulecolor=promptborder,
  frametitleaboveskip=6pt,
  frametitlebelowskip=6pt,
]{promptbox}

\AddToHook{cmd/appendix/before}{%
    \crefalias{section}{appendix}%
    \crefalias{subsection}{appendix}
}

\newcommand{\ulnum}[1]{%
  \sbox{\mathbox}{\num{#1}}\underline{\usebox{\mathbox}}%
}

\newcommand{\tablefont}{\fontsize{8pt}{10pt}\selectfont}
\newcommand{\OurMethod}{\textsc{OAK}\xspace}
\newcommand{\OurMethodCorrect}{\textsc{OAK+MEND}\xspace}

\newcommand{\OurInContext}{\textsc{In-Context}\xspace}
\newcommand{\domain}{\ensuremath{\mathsf{dom}}}
\newcommand{\range}{\ensuremath{\mathsf{rng}}}
\newcommand{\qualifiers}{\ensuremath{\mathsf{qual}}}
\let\oldtextsf\textsf
\renewcommand{\textsf}[1]{\oldtextsf{\fontsize{10}{10}\selectfont #1}}

\newcommand{\revision}[1]{{#1}}

\title{Better Later Than Sooner: Neuro-Symbolic Knowledge Graph Construction via Ontology-grounded Post-extraction Correction}

\author{
 \textbf{Lorenzo Loconte\textsuperscript{1,{\normalsize\dag}}} \\ l.loconte@sms.ed.ac.uk \\\And
 \textbf{Timothy Hospedales\textsuperscript{1,2}} \\ t.hospedales@samsung.com \\\And
 \textbf{Cristina Cornelio\textsuperscript{2}} \\ \href{c.cornelio@samsung.com}{c.cornelio@samsung.com} \\\AND\\[-32pt]
 \normalfont\textsuperscript{1}University of Edinburgh, UK \qquad
 \normalfont\textsuperscript{2}Samsung AI, Cambridge, UK
}%

\begin{document}
\maketitle

\begin{abstract}
Question answering (QA)
is a core
challenge in AI, particularly for complex queries requiring multi-hop reasoning across documents, or symbolic operations like aggregation or exhaustive listing.
Retrieval-augmented generation has become the dominant approach to QA, with recent graph-based variants addressing part of these issues by organizing
knowledge
to better support compositional questions.
However, most
\emph{textual}
graph-based RAG methods still lack the structure needed for symbolic operations
useful to answer complex questions reliably.  %
This motivates \emph{symbolic} graph-based approaches, which extract knowledge graphs (KGs) whose relations are logic predicates that
enable
SQL-like querying.
Yet these pipelines typically
use
LLMs for
KG extraction,
which can introduce consistency issues, where extracted facts may violate commonsense ontology constraints.
We propose a neuro-symbolic framework for ontology-grounded KG construction combining open-domain extraction, embedding-based canonicalization of types and predicates, and targeted LLM-based correction of ontology violations.
By deferring corrections to a post-extraction stage,
our method avoids repeated LLM calls, substantially reducing token usage while improving KG consistency and preserving downstream QA quality.
Finally, we show that the extracted
KGs are well suited
for symbolic querying by measuring the occurrence of SPARQL
graph patterns.
\end{abstract}

\makeatletter
\def\@makefnmark{} %
\makeatother
\footnotetext{\textsuperscript{\normalsize\dag}The work was done during an internship at the Samsung AI Center, Cambridge (UK). Correspondence to \href{c.cornelio@samsung.com}{c.cornelio@samsung.com}.}

\begin{figure}[t]
    \centering%
    \includegraphics[width=0.95\linewidth]{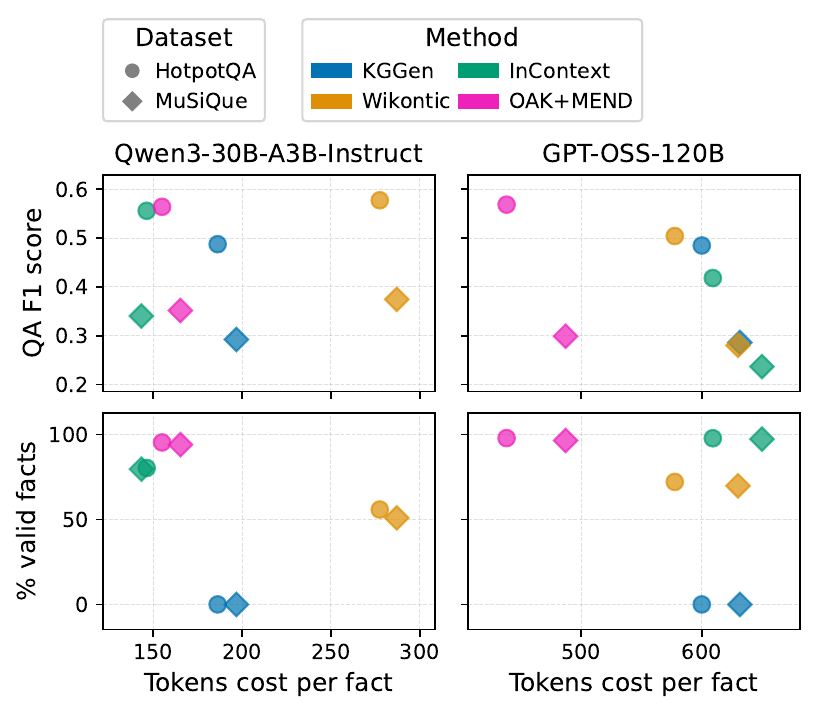}
    \caption{\textbf{Our \OurMethodCorrect method for ontology-based KG extraction achieves better trade-offs} regarding capturing the \emph{text semantics} (measured as QA performances, above row) and \emph{overall ontology consistency} (measured as \% of triples and qualifiers satisfying the ontology, below row), while being more efficient in terms of tokens per extracted fact (either triple or qualifier).
    We compare against the ontology-free \textsc{KGGen} \citep{mo2025kggen}, \textsc{Wikontic} \citep{chepurova2026wikontic} that iteratively aligns each triple to the ontology, and a baseline based on inserting the ontology as part of the LLM context during KG extraction (\OurInContext).}
    \label{fig:trade-off}
\end{figure}

\section{Introduction}\label{sec:introduction}

An important challenge in AI is answering user questions based on new or domain-specific information, often provided as collections of text documents \citep{voorhees2000trec,rajpurkar2016squad,lewis2020rag}.
The most popular method for question answering (QA) is retrieval-augmented generation (RAG), which decomposes the task into two steps: retrieving text fragments relevant for a given question and using the retrieved information to generate an answer \citep{chen2017reading,lewis2020rag}.
However, text-only QA methods often struggle with complex questions requiring multi-hop reasoning on several entities \citep{yang2018hotpotqa,trivedi2022musique}, symbolic operations such as arithmetic or aggregation (e.g., returning the answer that maximizes a quantity) \citep{dua2019drop,chen2021finqa,zhu2021tatqa}, or the composition of multiple intermediate answers
\citep{ho2020constructing}.
The challenge of answering these questions 
is further exacerbated by the fact that retrieving the relevant texts becomes increasingly hard as we scale the size of the corpora 
\citep{xiong2021approximate,weller2026theoretical}.

To address these problems, recent works have proposed 
structured indexes, typically graphs, which better capture global connections between documents and entities supporting multi-hop and compositional reasoning 
\citep{sarthi2024raptor,gutierrez2024hipporag,gutierrez2025from,edge2025localglobalgraphrag} and also enable agentic approaches balancing exploration and exploitation on the graph index directly \citep{gao2026beyond,du2026arag}.
Among these approaches, knowledge graphs (KG) are the most prominent, encoding direct relationships between entities as subject-predicate-object triples \citep{peng2023knowledge,pan2023large}.
KGs offer a formal symbolic relational structure enabling efficient traversal as required by multi-hop questions \citep{ren2020query2box,arakelyan2021complex-qa-tnorms,galkin2024foundation,gregucci2025complex}.

KGs often come with
an \emph{ontology}, which specifies a domain-specific schema, including predicates, entity types, and constraints on which type of entities each predicate can link.
This ontology enables verifying the consistency of the encoded information, and can guide the correction of constraint violations \citep{ahmetaj2022repairing,ferranti2024formalizing,lin2025systematic}.
Most importantly, it also makes KGs amenable to efficient retrieval through SQL-like languages such as SPARQL \citep{w3sparql}, which supports arithmetic and aggregation operations useful to answer challenging questions \citep{gashkov2025sparql,perevalov2025text-to-sparql}.

Recent works have shown that LLMs can effectively extract KGs from text \citep{zhu2023llms,mo2025kggen}.
However, extracting KGs that comply with an existing ontology remains challenging, as it requires balancing expressivity 
for QA, constraints satisfaction, and tokens usage.
One line of work addresses this by including ontology constraints
in the LLM context during extraction
\citep{mihindukulasooriya2023text2kgbench,vancauter2024ontology,nie2024knowledge,wang2025oskgc}.
However, for large ontologies such as Wikidata \citep{vrandeco2014wikidata}, this introduces a substantial computational overhead due to the required long context size.
Moreover, the consistency of the extracted KG ultimately depends on the ability of the chosen LLM to enforce the provided constraints, which makes smaller open LLMs perform worse.
Although we can leverage ad-hoc models to retrieve relevant ontology fragments for each text, these must be trained for the specific ontology considered \citep{zhang2024extract}.
Another line of work employs LLMs to extract open-domain triples, which are then iteratively aligned to the ontology using multiple LLM calls \citep{arun2025finreflectkg,lu2025karma,chepurova2026wikontic}.
While this offers flexibility in encoding relevant information, it requires many LLM calls per extracted triple, leading to a significant cost in terms of
tokens.

\textbf{Contributions.}
\textbf{(i)} We propose an efficient method for KG construction from text under ontology constraints that does not insert the constraints into the LLM context, but rather exploits text embeddings to map the extracted entity types and predicates to the ontology.
\textbf{(ii)} We symbolically detect the triples and qualifiers that violate the ontology and selectively correct each of them by prompting an LLM
with a candidate set of corrective actions.
This allows us to efficiently correct one or many triples (or qualifiers) with a single LLM call.
\textbf{(iii)} Our experiments show our method to be more token efficient than existing baselines, yet achieving ontology consistency rates for triples and qualifiers of up to 98.4\% and 96.8\% respectively.
QA experiments further show that our post-extraction corrections preserves crucial information from the text.
Then, to assess how amenable the extracted KGs are to symbolic querying via SPARQL, \textbf{(iv)} we introduce a benchmark based on graph patterns.

Code and data are available at \url{https://github.com/corneliocristina/OAK_MEND}.

\begin{figure*}[t]
    \centering%
    \includegraphics[width=0.875\linewidth]{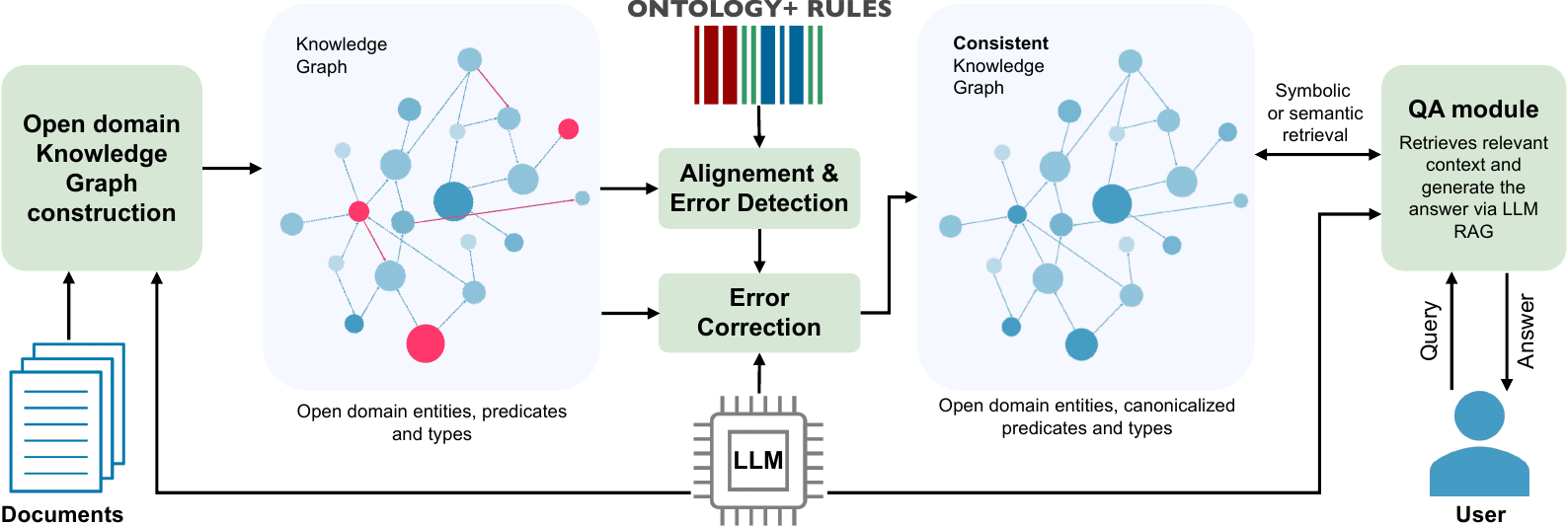}
\caption{\textbf{Overview of the \OurMethodCorrect method.} Documents are processed by an LLM to construct an open-domain knowledge graph, then aligned to the ontology by canonicalizing entity types and predicates. Symbolic rules derived from the ontology detect domain-range violations (both for triples and qualifiers). Detected inconsistencies are corrected through LLM calls, producing a consistent KG for (symbolic/semantic) retrieval for, e.g., QA tasks.}
\label{fig:okgc_overview}
\end{figure*}

\section{Background: Ontology Constraints}\label{sec:background-ontology-constraints}

A KG $\calG$ is a directed graph where links are encoded using subject-predicate-object triples: $\calG=\{(s_i,r_i,o_i)\}_i\subset\calE\times\calR\times\calE$ where $\calE$ and $\calR$ denote the sets of entities and predicates, respectively.
KGs may also include qualifiers useful to refine or contextualize the semantics of triples.
Given a set of qualifier predicates $\calQ\subset\calR$, a qualifier is a pair $(r_q,o_q)$, with $r_q\in\calQ$, $o_q\in\calE$, that is associated to a single triple.
For instance, the sentence ``Alan Turing earned his Ph.D. at Princeton University in 1938'' can be modeled as the triple $(\textsf{Alan Turing}, \textsf{educated at}, \textsf{Princeton Univ.})$ equipped with the qualifier $(\textsf{point in time},\textsf{1938})$.
A key component of a KG is its \emph{ontology}, which define the admissible entity types and predicates, as well as the constraints they must satisfy.
Following the Wikidata data model \citep{vrandeco2014wikidata}, we consider an ontology consisting of (1) a hierarchy of entity types, (2) domain-range constraints, and (3) constraints defined over qualifiers.
Below we formalize each ontology component.

\textbf{Entity types hierarchy.}
Let $\calT$ be a set of entity types, and let $t_1\prec t_2$ denote the ``is-subclass-of'' partial ordering between types $t_1,t_2\in\calT$.
Moreover, for each entity $e\in\calE$ we denote by $\tau(e)\subset\calT$ the set of its associated types.
For example, $\textsf{mathematician}\in\tau(\textsf{Alan Turing})$ states that \textsf{Alan Turing} is typed as \textsf{mathematician}, while $\textsf{mathematician}\prec\textsf{human}$ specifies that \textsf{mathematician} is a subclass of the type \textsf{human}.

\textbf{Domain-range constraints.}
For each predicate $r\in\calR$, let $\domain(r)\subseteq\calT$ and $\range(r)\subseteq\calT$ denote the \emph{domain} and \emph{range} constraints of $r$, respectively.
These specify the admissible subject and object types for triples involving  $r$.
A triple $(s,r,o)$ satisfies the domain constraint if $t_s\prec u_s$ for some $t_s\in\tau(s),u_s\in\domain(r)$,
and satisfies the range constraint if $t_o\prec u_o$ for some $t_o\in\tau(o),u_o\in\range(r)$.
For example, given
\vspace{-3pt}
\begin{align*}
    &\domain(\textsf{educated at}) = \{ \textsf{human} \}\\
    &\range(\textsf{educated at}) = \{ \textsf{university} \}, \text{then}\\[-20pt]
\end{align*}
$(\textsf{Alan Turing},\textsf{educated at},\textsf{Princeton University})$ 
satisfies the domain constraint of \textsf{educated at} since $\textsf{mathematician}\prec\textsf{human}$; and the range constraint whenever $\textsf{university}\in\tau(\textsf{Princeton University})$.

While prior work \citet{feng2024ontology,chepurova2026wikontic} studied KG extraction under domain-range constraints, here we 
additionally consider qualifiers constraints, which we formalize next.

\textbf{Qualifier constraints.}
Each predicate $r\in\calR$ specifies a set $\qualifiers(r)\subseteq\calQ\subset\calR$ of admissible qualifier predicates.
For example, the qualifier $(\textsf{point in time},\textsf{1938})$ from above is consistent with $\textsf{educated at}$ if $\textsf{point in time}\in\qualifiers(\textsf{educated at})$, 
whereas $(\textsf{color},\textsf{red})$ is inconsistent if $\textsf{color}\notin\qualifiers(\textsf{educated at})$.
Furthermore, a qualifier $(r,o)$ satisfies the range constraint of $r$ if $t_o\prec u_o$ for some $t_o\in\tau(o),u_o\in\range(r)$.

\textbf{Symbolic querying via SPARQL.}
The ontology constraints defined above provide the \emph{rules} necessary to write SQL-like queries for symbolic retrieval or QA, using SPARQL \citep{w3sparql}.
For example, consider the task of retrieving \emph{all physicists who received the Nobel prize after 2000}.
This can be expressed as the following query.
\vspace{-6pt}%
\begin{minted}[linenos,autogobble,numbersep=3pt,framesep=0pt,fontsize=\small]{sparql}
SELECT ?person WHERE {
    ?person instanceOf: Human: .
    ?person occupation: Physicist: .
    ?person awardReceived: ?stmt .
    ?stmt awardValue: NobelPrizePhysics: ;
          pointInTime: ?date .
    FILTER ( ?date >= "2000-01-01" ) }
\end{minted}
\vspace*{-6pt}
In the above, L2 specifies a type constraint over the entity variable \verb|?person|, while in L3 the range constraint of \verb|occupation:| tells us to put the occupation \verb|Physicist:| on the right-hand side of the triple.
The fact that \verb|pointInTime:| is an allowed qualifier by \verb|awardReceived:| guarantees we can use it to access the additional time information of the award relationship, as shown in L5-L7 using statement expressions following the Wikidata data model \citep{vrandeco2014wikidata}.

In \cref{sec:our-method} we propose a method for KG extraction that maps open-domain types and predicates to the ones allowed by the ontology.
Then, in \cref{sec:our-method-corrections-step} we develop an approach to systematically correct violations of ontology constraints.

\section{KG Extraction and Canonicalization}\label{sec:our-method}

Given a source text $\calS$, we aim at extracting a KG $\calG$ equipped with typed entities and qualifiers such that it encodes the relationships between entities in $\calS$.
Here, we present a method that (1) performs open-domain KG construction, and (2) maps (or canonicalizes) the extracted open-domain types (resp. predicates) to the types (resp. predicates) allowed by the ontology, i.e., $\calT$ (resp. $\calR$).
To make (2) efficient, we leverage embeddings computed on types and predicates labels, while we use LLMs for disambiguation only.
We dub this method as \emph{ontology-aligned KG construction} (\OurMethod), and defer the problem of satisfying the ontology constraints to \cref{sec:our-method-corrections-step}.
In \cref{app:prompts} we report all our prompts.

\subsection{Open-domain Extraction}\label{sec:extraction-step}

\textbf{Step 1.}
We use a single prompt instructing an LLM to extract triples, entity types, and optionally a number of qualifiers for each triple from the source text $\calS$.
All extracted elements are open-domain: the ontology is not part of the input context, and the extracted types and predicates are instead mapped to it in subsequent steps (see \cref{sec:canonicalization-step}).
This KG extraction step differs from the one in \citet{chepurova2026wikontic} by the fact that we also extract the types of qualifier objects, as required by the constraints over qualifiers (see \cref{sec:background-ontology-constraints}).
We use a fixed in-context example to improve the adherence to the required JSON output format.

Open-domain KG extraction avoids having to first identify which ontology fragments are relevant to the input text and include them in the LLM prompt, as done in \citet{zhang2024extract,feng2024ontology}.
As a result, the prompt is shorter, reducing token usage
and thereby improving prefix cache hit rates as well.
\revision{%
In addition, in our experiments using a reasoning LLM, we find that providing the ontology constraints as part of the input context drastically reduces the number of extracted triples (\cref{sec:empirical-evaluation-consistency}), which in turn reduces QA performances as well (\cref{sec:empirical-evaluation-qa}).
Perhaps surprising, this represents evidence that leveraging constraints at KG extraction time can sometimes negatively affect KG construction and QA.
}%

\subsection{Canonicalization of Types and Predicates}\label{sec:canonicalization-step}

\textbf{Step 2.} We map (or canonicalize) the open-domain types and predicates extracted in \textbf{Step 1} to types and predicates allowed by the ontology, respectively.
Recently, \citet{chepurova2026wikontic} proposed to \emph{perform multiple LLM calls for each extracted triple} with the purpose of canonicalizing the subject and object entity types, followed by canonicalizing the predicate such that the resulting triple satisfies the domain-range constraint.
However, this accounts for a significant tokens overhead when applied to several thousands triples.

We seek a more token efficient approach by making the following simplifying assumptions.
First, we assume that entities that have been associated to the same open-domain type in \textbf{Step 1} should have the same canonical type from the ontology.
Second, we assume that extracted triples sharing the same open-domain predicate express the same relationship semantics, and therefore their predicates should be canonicalized in the same way.
\revision{%
These simplifying assumptions are not expected to hold universally, and their purpose is to minimize the number of LLM calls at the cost of introducing schema violations.
Nevertheless, these violations can be later resolved in our framework based on post-extraction corrections (see \cref{sec:our-method-corrections-step}).
We exploit the assumptions above in a canonicalization step leveraging embedding similarities.
}%

Formally, let $\hat{t}$ be a type label that has been extracted in \textbf{Step 1}, which we aim to map to some $t\in\calT$.
We compute the text embedding similarity between the extracted $\hat{t}$ and the label of each $t\in\calT$, and pick the type having the highest similarity score.
Following \citet{chepurova2026wikontic} we also consider any available label aliases for each $t$.
However, multiple types might have very similar labels or even have common aliases, making canonicalization through embedding similarities alone insufficient.
Whenever multiple types have close similarity scores w.r.t. $\hat{t}$ based on a
hyperparameter
$\beta$, we prompt an LLM to select the most suitable one.
We construct the set of candidates types as
\vspace{-3pt}
\begin{equation*}
    \calC_{\calT}(\hat{t}) = \{t_i\in\calT\mid \phi(\hat{t},t_i) \geq m_{\hat{t}} - \beta\} \vspace*{-3pt}
\end{equation*}
where $m_{\hat{t}} = \max_{t_i\in\calT} \{ \phi(\hat{t},t_i) \}$ and $\phi(\hat{t},t_i)$ denotes the maximum embedding similarity between $\hat{t}$ and any label or alias of $t_i$.
Then, the canonical $t$ for $\hat{t}$ is chosen by calling an LLM, i.e., $t = \mathrm{LLM}(\calS,\hat{t},\calC_{\calT}(\hat{t}),E_{\hat{t}})$, where $E_{\hat{t}}$ denotes the set of entities that have been previously typed with $\hat{t}$.
In the particular case of $|\calC_{\calT}(\hat{t})|=1$, no LLM calls are made and thus $\beta$ can be used to tune the tokens cost.
In our experiments we use the cosine similarity and empirically set $\beta=0.05$.

We proceed similarly to canonicalize the predicates.
Given an open-domain predicate label $\hat{r}$ extracted in \textbf{Step 1}, we build the set of candidates
\vspace{-3pt}
\begin{equation*}
    \calC_{\calR}(\hat{r}) = \{r_i\in\calR\mid \phi(\hat{r},r_i) \geq m_{\hat{r}} - \beta\}, \vspace*{-3pt}
\end{equation*}
where $m_{\hat{r}} = \max_{r_i\in\calR} \{ \phi(\hat{r},r_i)\}$.
Then, the canonical $r\in\calR$ for $\hat{r}$ is chosen by calling an LLM, i.e., $r = \mathrm{LLM}(\calS,\hat{r},\calC_{\calR}(\hat{r}),T_{\hat{r}})$, where $T_{\hat{r}}\subset\calE\times\calE$ denotes the subject-object entity pairs that have been linked with the open-domain predicate $\hat{r}$ in \textbf{Step 1}.

Finally, since qualifier predicates specify additional information about the triples themselves, their number is typical much smaller than the total number of predicates (e.g., it is only \textasciitilde29\% in the Wikidata ontology fragment we consider in our experiments).
This reduces ambiguity during canonicalization, 
allowing us to simply map each qualifier predicate name to
the higher scoring qualifier predicate in the ontology.
Finally, we remove duplicate entitites as described next.

\subsection{Types-Aware Entity Deduplication}\label{sec:entity-dedup-step}

\textbf{Step 3.}
A common approach to extracting a KG from a corpus consists of splitting the texts into
chunks,
running the KG extraction method on each
chunk
and then merging the obtained KGs \citep{zhang2024extract,mo2025kggen}.
However, this often yields entities with different labels that refer to the same concept, making an entity deduplication step necessary.
Recently, \citet{mo2025kggen} proposed a two-step algorithm 
that first clusters entities based on their embeddings,
and then uses an LLM to iteratively identify duplicates within each cluster.
We adopt this method while also exploiting the type information from \textbf{Step 2}.
In particular, entities typed as literals (e.g., a number or a calendar date)
are excluded from
deduplication, hence reducing the clusters size.

Until now, we discussed \OurMethod: a method to extract a KG such that entity types and predicates are mapped to the ones in a given ontology.
However, the extracted KG might still contain ontology constraint violations, which we aim to correct next.

\begin{table*}[!t]
    \centering%
    \tablefont%
    \setlength{\tabcolsep}{6pt}%
    \begin{tabular}{%
    l
    S[table-format=6(4), separate-uncertainty=true]
    S[table-format=2.1(1.1), separate-uncertainty=true]
    S[table-format=5(3), separate-uncertainty=true]
    S[table-format=2.1(1.1), separate-uncertainty=true]
    S[table-format=3.1]
    S[table-format=2.1]
    S[table-format=2.1]
    @{~(} 
    S[table-format=1.2]
    @{)}
    }
    \toprule
    \textbf{Dataset: HotpotQA} &
    \multicolumn{2}{c}{\textbf{Triples Statistics}} &
    \multicolumn{2}{c}{\textbf{Qualifiers Statistics}} &
    \multicolumn{3}{c}{\textbf{\# Tokens $(\times 10^6)$}} \\
    & {\# Total} & {\% Valid} & {\# Total} & {\% Valid} & {Prompt} & {Completion} & \multicolumn{2}{c}{Weighted} \\
    \midrule
    {\cellcolor{lightgray} \textsc{Qwen3-30B-A3B}} \\
    \textsc{KGGen}
        & 155391 \pm  185 &        {---} &       0 \pm 0 &        {---} &  71.7 &  11.0 & 28.9 & 0.64 \\
    \textsc{Wikontic}
        & 117891 \pm   79 & 77.4 \pm 0.2 & 45438 \pm 455 &        {---} & 139.3 &  10.5 & 45.3 & 1.00 \\
    \OurInContext
        & 173942 \pm  420 & {\ulnum{79.4 \pm 0.2}} & 70763 \pm 352 & {\ulnum{82.6 \pm 0.2}} &  93.3 &  12.5 & 35.8 & 0.79 \\
    \OurMethod
        & 127763 \pm  107 & 63.4 \pm 0.1 & 45721 \pm 516 & 33.6 \pm 0.1 &  34.3 &   7.7 & \bfseries 16.3 & 0.36 \\
    \OurMethodCorrect
        & 127108 \pm  130 & \bfseries 96.8 \pm 0.1 & 45721 \pm 516 & \bfseries 91.5 \pm 0.1 &  74.0 &   8.3 & {\ulnum{26.8}} & 0.59 \\
    \midrule
    {\cellcolor{lightgray} \textsc{GPT-OSS-120B}} \\
    \textsc{KGGen}
        & 128163 \pm 4440 &        {---} &     0 \pm   0 &        {---} &   87.8 &  54.8 & 76.8 & 0.88 \\
    \textsc{Wikontic}
        & 110109 \pm  329 & \bfseries 99.0 \pm 0.1 & 40939 \pm  35 &        {---} &  139.5 &  52.4 & 87.3 & 1.00 \\
    \OurInContext
        &  67197 \pm  144 & 97.9 \pm 0.1 &  8639 \pm 455 & \bfseries 97.7 \pm 0.1 &   98.9 &  21.5 & {\ulnum{46.2}} & 0.53 \\
    \OurMethod
        & 113843 \pm  237 & 67.9 \pm 0.3 & 45351 \pm  26 & 35.7 \pm 0.3 &   33.8 &  28.4 & \bfseries 36.8 & 0.42 \\
    \OurMethodCorrect
        & 112737 \pm 1093 & {\ulnum{98.4 \pm 1.0}} & 45176 \pm 273 & {\ulnum{96.8 \pm 1.7}} &   67.7 &  52.3 & 69.2 & 0.79 \\
    \bottomrule
    \end{tabular}
    \caption{\textbf{KG Construction with \textsc{\OurMethodCorrect} achieves high ontology consistency rates for both triples and qualifiers, yet requires modest token cost. }%
We show the number of triples and qualifiers extracted by different methods for KG construction, as well as the percentage consistent with ontology constraints (see \cref{sec:background-ontology-constraints}), where applicable.
We also show the total number of prompt and completion tokens (in millions), and a weighted cost combination that takes into account a prompt-to-completion cost ratio of $\sfrac{1}{4}$ (e.g., as the \$ price ratio for GPT-4.1).}
    \label{tab:empirical-evaluation-kgc-stats-hotpotqa}
\end{table*}

\section{Correction of Ontology Violations}\label{sec:our-method-corrections-step}

The entity types allow us to symbolically detect which triples and qualifiers violate which ontology constraints.
This contrasts with methods that rely on LLMs for error detection, such as detecting disallowed types \citep{arun2025finreflectkg}, classifying class membership errors \citep{allen2024evaluating}, and supporting human-in-the-loop validation \citep{regino2025can}.

After we collect the triples and qualifiers violating the ontology, we prompt an LLM to correct them by choosing from a set of syntactic transformations.
Note that, since we correct ontology violations \emph{after} the KG has been extracted, our method can be applied \emph{ex-post}
to any KG extraction approach,
Next, we describe our approach to correct inconsistent triples.

\subsection{Correcting Triples}\label{sec:our-method-corrections-step-correct-triples}

To motivate the set of syntactic transformations useful to correct ontology violations, we identify a number of KG extraction mistakes -- each of which suggesting a particular corrective action.

\noindent%
\textbf{V1.}
The \textbf{subject-object inversion} error occurs when the types of the subject $s$ (resp. object $o$) in a triple $(s,r,o)$ are inconsistent with the domain (resp. range) of the predicate $r$ 
because the triple direction is reversed, while the inverted triple 
$(o,r,s)$ is instead consistent.
For example, if $\domain(\textsf{author})=\{\textsf{book}\}$ and $\range(\textsf{author}) = \{\textsf{human}\}$, then the triple $(\textsf{G. Orwell},\textsf{author},\textsf{Animal Farm})$ is inconsistent when $\tau(\textsf{G. Orwell}) = \{\textsf{human}\}$ and $\tau(\textsf{Animal Farm}) = \{\textsf{book}\}$.
Although this triple expresses that \textsf{G. Orwell} is the author of \textsf{Animal Farm}, 
it violates the ontology because the subject and object are reversed.
It can therefore be corrected by
\emph{swapping subject and object}.
We find this mistake to be frequent, and thus automatically apply the correction whenever possible.

\noindent%
\textbf{V2.} The error of \textbf{predicate mis-canonicalization} occurs when a predicate has been wrongly selected:
the triple
$(s,r,o)$ is inconsistent because $r$ does not explain a relationship between
entities
$s$ and $o$.
For example, the triple $(\textsf{Animal Farm},\textsf{distribution format},\textsf{Penguin Books})$ might have been hallucinated because the publisher \textsf{Penguin Books} has ``Books'' in its label.
This triple is inconsistent because the type \textsf{organization} of \textsf{Penguin Books} is not allowed by the range constraint of \textsf{distribution format}.
Instead, $(\textsf{Animal Farm},\textsf{publisher},\textsf{Penguin Books})$ 
satisfies the domain-range constraint of \textsf{publisher}.
Thus, the corrective action is to \emph{replace the predicate}.

\noindent%
\textbf{V3.}
Lastly, a triple $(s,r,o)$ may be inconsistent because of \textbf{underspecified or missing types}: 
the types of the subject $s$ (resp. the object $o$) are not specific enough to satisfy the domain (resp. the range) of $r$ or, alternatively, a type
might be
missing.
For example, assuming that $\tau(\textsf{Animal Farm}) = \{\textsf{written work}\}$, then $(\textsf{Animal Farm},\textsf{author},\textsf{G. Orwell})$ is inconsistent with the domain of $\textsf{author}$, as $\textsf{book}\prec\textsf{written work}$.
That is, the type \textsf{written work} is not specific enough as required by the predicate \textsf{author}.
Here, the correction is \emph{adding a type to the subject}.

The cases \textbf{V1-3} above motivate prompting an LLM to choose among the corrective actions:
\begin{itemize}[noitemsep,nosep]
    \item \textsf{swap}: swap the subject $s$ and object $o$ entities.
    \item \textsf{replace\_predicate}: substitute $r$ with $r'\in\calR$ such that $(s,r',o)$ is consistent. The candidates $r'$ are ranked based on the embedding similarity w.r.t. the label of $r$.
    \item \textsf{add\_subject\_type}: add a new type to $s$, where the candidate types are in $\domain(r)$.
    \item \textsf{add\_object\_type}: add a new type to $o$, where the candidate types are in $\range(r)$.
\end{itemize}
Since multiple actions might be needed to fix a domain-range violation (e.g., adding a type followed by swapping), the LLM is prompted to return a composition of actions.
Moreover, by allowing the LLM to add entity types and by propagating the updated types to all triples, \emph{our method can effectively correct multiple constraint violations with a single LLM call}, making it particularly efficient.

\begin{table*}[!t]
    \tablefont%
    \centering%
    \setlength{\tabcolsep}{6pt}%
    \begin{tabular}{%
    l
    S[table-format=2.1(1.1), separate-uncertainty=true]
    S[table-format=2.1(1.1), separate-uncertainty=true]
    S[table-format=2.1(1.1), separate-uncertainty=true]
    S[table-format=2.1(1.1), separate-uncertainty=true]
    S[table-format=2.1(1.1), separate-uncertainty=true]
    S[table-format=2.1(1.1), separate-uncertainty=true]
    S[table-format=2.1(1.1), separate-uncertainty=true]
    S[table-format=2.1(1.1), separate-uncertainty=true]}
    \toprule
    {\multirow{3}{*}{\textbf{Method}}} &
    \multicolumn{4}{c}{\textbf{HotpotQA}} &
    \multicolumn{4}{c}{\textbf{MuSiQue}} \\
    & \multicolumn{2}{c}{\cellcolor{lightgray} \textsc{Qwen3-30B-A3B}} & \multicolumn{2}{c}{\cellcolor{lightgray} \textsc{GPT-OSS-120B}} & \multicolumn{2}{c}{\cellcolor{lightgray} \textsc{Qwen3-30B-A3B}} & \multicolumn{2}{c}{\cellcolor{lightgray} \textsc{GPT-OSS-120B}} \\
    & {{EM}} & {{F\textsubscript{1}}} & {{EM}} & {{F\textsubscript{1}}} & {{EM}} & {{F\textsubscript{1}}} & {{EM}} & {{F\textsubscript{1}}} \\
    \midrule
    \textsc{ZeroShot}
        & 21.8 \pm 0.3 & 30.7 \pm 0.3 & 33.4 \pm 0.7 & 45.1 \pm 0.5
        &  2.3 \pm 0.1 &  9.4 \pm 0.7 & 12.9 \pm 0.7 & 24.1 \pm 0.2 \\
    \textsc{VectorRAG}
        & 42.9 \pm 0.4 & 56.5 \pm 0.3 & 58.6 \pm 0.2 & 72.9 \pm 0.1
        & 14.0 \pm 0.4 & 23.2 \pm 0.5 & 30.4 \pm 0.7 & 43.6 \pm 0.8 \\
    \textsc{HippoRAGv2}
        & 50.0 \pm 0.8 & 64.1 \pm 0.6 & 38.8 \pm 0.7 & 57.0 \pm 0.5  
        & 22.0 \pm 0.1 & 33.9 \pm 1.0 & 19.4 \pm 0.1 & 36.2 \pm 0.2 \\
    \midrule
    \textsc{KGGen}
        & 38.5 \pm 1.5 & 48.7 \pm 1.6 & 38.9 \pm 0.5 & 48.5 \pm 1.3
        & 19.2 \pm 0.2 & 29.2 \pm 0.2 & {\ulnum{21.9 \pm 0.6}} & 28.6 \pm 0.6 \\
    \textsc{Wikontic}
        & {\ulnum{46.4 \pm 0.7}} & {\ulnum{57.7 \pm 0.5}} & 40.4 \pm 0.8 & 50.4 \pm 0.8
        & \bfseries 25.0 \pm 1.0 & \bfseries 37.4 \pm 0.1 & 19.9 \pm 0.2 & 28.0 \pm 0.6 \\
    \OurInContext
        & 45.0 \pm 0.7 & 55.6 \pm 0.9 & 33.1 \pm 0.7 & 41.8 \pm 0.8
        & 22.0 \pm 0.7 & 34.0 \pm 0.8 & 17.5 \pm 0.5 & 23.7 \pm 0.7 \\
    \OurMethod
        & \bfseries 47.0 \pm 1.0 & \bfseries 58.6 \pm 0.7 & \bfseries 46.4 \pm 0.9 & \bfseries 57.8 \pm 1.2
        & {\ulnum{24.4 \pm 0.7}} & {\ulnum{36.5 \pm 0.9}} & \bfseries 22.6 \pm 1.3 & \bfseries 31.1 \pm 0.7 \\
    \OurMethodCorrect
        & 45.6 \pm 0.4 & 56.4 \pm 0.5 & {\ulnum{46.1 \pm 2.0}} & {\ulnum{56.8 \pm 1.7}}
        & 23.3 \pm 1.1 & 35.2 \pm 1.4 & 21.5 \pm 0.1 & {\ulnum{29.9 \pm 0.2}} \\
    \bottomrule
    \end{tabular}
\vspace{-1pt}
\caption{\textbf{\OurMethod achieves competitive QA results, and our corrections enhancing ontology consistency account for little or no loss of performances.}
We show QA performances in terms of exact match (EM) and $\mathrm{F}_1$ metrics computed on predicted and ground-truth answers, achieved by leveraging the multi-step QA method originally developed in \citet{chepurova2026wikontic}. For \textsc{HippoRAGv2} we report QA results achieved by running their code.%
}%
    \label{tab:empirical-evaluation-qa}
\end{table*}

\subsection{Correcting Qualifiers}\label{sec:our-method-corrections-step-correct-qualifiers}

To correct qualifiers violating the ontology, we proceed similarly to the triples corrections above.
As detailed in \cref{app:detailed-corrections}, qualifier constraints indicate two main sources of errors:
using a disallowed qualifier predicate, 
or the types of the qualifier object are underspecified/missing (see \cref{sec:background-ontology-constraints}).
Given an inconsistent qualifier $(r_q,o_q)$ associated to a triple $(s,r,o)$, we prompt an LLM to choose one corrective action among:
\begin{itemize}[noitemsep,nosep]
    \item \textsf{replace\_predicate}: substitute $r_q$ with $r_q'\in\textsf{qual}(r)$ such that $(r_q',o_q)$ is consistent. The candidates $r_q'$ are ranked based on the embedding similarity w.r.t. the label of $r_q$.
    \item \textsf{add\_object\_type}: add a new type to $o_q$, where the candidate types are in $\range(r_q)$.
\end{itemize}

In all correction prompts we insert a one-line explanation of why a triple or qualifier violates a constraint.
To keep corrections grounded in the source text, we also
add the text fragment from which the inconsistent triple or qualifier was originally extracted from.
Finally, we give the option to \emph{not} choose any action whenever 
all possible corrections would yield a fact unsupported by the text.

\subsection{\revision{Corrective Actions Re-validation}}

\revision{Each corrective action is followed by ontology-based re-validation. Once a triple or qualifier is deemed consistent with the ontology, it is retained and excluded from subsequent corrections, preventing the introduction of new ontology violations. MEND performs up to two correction iterations, as some facts require sequential actions (e.g., adding a type and then swapping the entities). The remaining invalid facts correspond to cases in which the LLM fails to identify an action that produces an ontology-valid result while preserving consistency with the input text. We observed no significant improvement in the considered metrics when increasing the number of iterations beyond two.}

\section{Empirical Evaluation}\label{sec:empirical-evaluation}

We answer the following questions:
\textbf{Q1.} How consistent are the KGs extracted by our method w.r.t. an ontology, and how token efficient it is?
\textbf{Q2.} Does our algorithm correcting ontology violations still preserve the text semantics as required by the QA task?
\textbf{Q3.} Are the extracted KGs well suited for symbolic querying via SPARQL queries?

\textbf{Datasets and ontology.}
Following \citet{gutierrez2024hipporag,chepurova2026wikontic}, we focus on fragments of HotpotQA \citep{yang2018hotpotqa} and MuSiQue \citep{trivedi2022musique} datasets consisting of text paragraphs and questions in natural language.
Following \citet{feng2024ontology,chepurova2026wikontic}, we also use a Wikidata ontology fragment, consisting of a types hierarchy, allowed predicates, and the related constraints \citep{vrandeco2014wikidata}.
We detail these in \cref{app:datasets-ontology}.

\textbf{Baselines.}
We compare against: \textsc{KGGen}, an ontology-free KG extraction method \citep{mo2025kggen}; \textsc{Wikontic}
targeting
the Wikidata type hierarchy and domain-range constraints \citep{chepurova2026wikontic}; and a custom-built in-context baseline inserting a verbalization of the constraints defined in \cref{sec:background-ontology-constraints} as part of the LLM
context (details in \cref{app:in-context-baseline}).
Although \textsc{KGGen} does not rely on an existing ontology, we still compare against it in terms of efficiency and downstream QA performances.
To the best of our knowledge, \textsc{Wikontic} is the closest to our method with an open implementation.
Following the QA setting adopted by \textsc{Wikontic}, we retain triples and qualifiers violating the ontology.
\revision{%
We also report results of QA methods retrieving source text fragments and using them as part of the LLM context when answering questions, such as vector RAG \citep{lewis2020rag} and HippoRAGv2 \citep{gutierrez2025from}.
}%

\textbf{LLMs.}
All our experiments use either the instruction-tuned \href{https://huggingface.co/Qwen/Qwen3-30B-A3B-Instruct-2507-FP8}{\textsc{Qwen3-30B-A3B-Instruct-2507-FP8}} (\textsc{Qwen3-30B-A3B} for brevity) or the reasoning model \href{https://huggingface.co/openai/gpt-oss-120b}{\textsc{GPT-OSS-120B}}, together with 
 \href{https://huggingface.co/Qwen/Qwen3-Embedding-0.6B}{\textsc{Qwen3-Embedding-0.6B}} for text embeddings.

\subsection{Q1. Consistency and Efficiency}\label{sec:empirical-evaluation-consistency}

In \cref{tab:empirical-evaluation-kgc-stats-hotpotqa} we show
the number of triples and qualifiers extracted on HotpotQA, as well as the fraction of triples and qualifiers that are consistent.
By applying our post-extraction corrections (\cref{sec:our-method-corrections-step}), dubbed as \OurMethodCorrect,
we improve the ontology consistency of both triples and qualifiers extracted by \OurMethod (\cref{sec:our-method})
i.e.,
we increase
the percentage of valid triples (resp. qualifiers) from 63.4\% (resp. 33.6\%) to 96.8\% (resp. 91.5\%) with \textsc{Qwen3-30B-A3B}; and from 67.9\% (resp. 35.7\%) to 98.4\% (resp. 96.8\%) with \textsc{GPT-OSS-120B}.
When using \textsc{Qwen3-30B-A3B}, the achieved ontology consistency is higher than the \OurInContext baseline and \textsc{Wikontic}, while still comparable when using \textsc{GPT-OSS-120B}.
Yet our \textsc{\OurMethodCorrect} method is more efficient in terms of prompt and completion tokens.
We show this
by weighting prompt and completion tokens differently given their different
overhead.
When compared to the token cost of \textsc{Wikontic} (normalized 100\%), the cost of \textsc{\OurMethodCorrect} is 59\% on \textsc{Qwen3-30B-A3B}; and 79\% on \textsc{GPT-OSS-120B}.
\cref{tab:empirical-evaluation-kgc-stats-musique} shows similar results on MuSiQue.

\revision{%
As a minor observation,
we note a slight decrease in the number of triples from \OurMethod to \OurMethodCorrect in \cref{tab:empirical-evaluation-kgc-stats-hotpotqa}.
This decrease is due to deduplication: because triples are stored as a set, correcting an inconsistent triple may make it identical to an existing one, causing the duplicate to be automatically removed.
}%
\revision{%
Furthermore, \cref{tab:corrective-actions} reports the frequency of each triple corrective action kind as predicted by the LLMs on HotpotQA, showing that predicate replacement is the action predicted with the highest frequency.
}%

\revision{%
Instead,
we do not report a ``no action'' rate because ``no action'' can result from several situations, making this rate difficult to compare meaningfully. For example, cases in which the LLM selects a replacement predicate outside the candidate list are also considered ``no action.'' Nevertheless, given MEND’s near-100\% consistency rate, such cases are expected to be relatively small when compared to the reported action frequencies.
}%

\begin{table}[!h]
\tablefont%
\setlength{\tabcolsep}{5pt}%
\begin{tabular}{lS[table-format=5(3), separate-uncertainty=true]S[table-format=5(4), separate-uncertainty=true]}
\toprule
 & \textsc{Qwen3-30B-A3B}     & \textsc{GPT-OSS-120B}     \\
\midrule
\textbf{Triples}                     &                            &                           \\
\texttt{swap$^*$}               &  9671 \pm 35               &  8520 \pm 1175            \\
\texttt{add\_subject\_type} &  1119 \pm 30               &  3148 \pm  775            \\
\texttt{add\_object\_type}  &   694 \pm 53               &  5784 \pm  459            \\
\texttt{replace\_predicate} & 45539 \pm 30               & 34715 \pm  503            \\
\textbf{Qualifiers}                  &                            &                           \\
\texttt{add\_object\_type}  &  1678 \pm 9                &  2132 \pm  354            \\
\texttt{replace\_predicate} & 11581 \pm 228              & 12777 \pm  259            \\
\bottomrule
\end{tabular}
\caption{\revision{%
\textbf{Predicate replacement and subject-object swap are the most common actions.} We report the frequency of predicted corrective actions for triples and qualifiers.
When using \textsc{Qwen3-30B-A3B} (resp. \textsc{GPT-OSS-120B}), the number of triples whose either subject or object types have been updated is $12058.7 \pm 246.4$ (resp. $27408.7 \pm 571.8$),
hence showing that the impact of subject and object types addition on the triples corrections is still lower than the replacement of predicates.
$^*${\footnotesize\emph{Swap} statistics include both modalities: LLM-predicted swaps and automatic swaps when they yield a valid triple.}
}}%
\label{tab:corrective-actions}
\end{table}

\revision{%
\textbf{Qualitative evaluation.}
Our framework alters the extracted entity types and triples, which can in turn modify the encoded semantics.
In order to rule out whether \OurMethodCorrect fixes the ontology violations by sacrificing semantic faithfulness, \cref{app:human-qualitative-evaluation} reports a human audit measuring the alignment of extracted triples and entity types w.r.t. the source text \emph{before} and \emph{after} corrections. 
After the triples corrections, \cref{tab:human-qualitative-evaluation} shows an increase in triples faithfulness and no loss in types faithfulness on HotpotQA.
Although brief, our analysis suggests that the ontology consistency improvements shown in \cref{tab:empirical-evaluation-kgc-stats-hotpotqa} do not come at the cost of a lower text fidelity.
In the next section, this claim is further validated quantitatively by evaluating \OurMethodCorrect on downstream QA.
}%

\begin{table*}[!t]
    \centering%
    \tablefont%
    \setlength{\tabcolsep}{4.75pt}%
    \begin{tabular}{%
    l
    S[table-format=1.2, separate-uncertainty=true]
    S[table-format=2.1, separate-uncertainty=true]
    S[table-format=3.1(1.1), separate-uncertainty=true]
    S[table-format=4.1(2.1), separate-uncertainty=true]
    S[table-format=1.2, separate-uncertainty=true]
    S[table-format=1.2, separate-uncertainty=true]
    S[table-format=3.1(1.1), separate-uncertainty=true]
    S[table-format=4.1(2.1), separate-uncertainty=true]
    }
    \toprule
    {\multirow{3}{*}{\textbf{Method}}} &
    \multicolumn{4}{c}{\textbf{HotpotQA}} &
    \multicolumn{4}{c}{\textbf{MuSiQue}} \\
    \cmidrule[0.5pt](lr{0.3em}){2-5} \cmidrule[0.5pt](lr{0.3em}){6-9} %
    & {Avg. edge} & {\%$(s, o)$ w/} & {\multirow{2}{*}{h-index}} & {\multirow{2}{*}{i100-index}} & {Avg. edge} & {\%$(s, o)$ w/} & {\multirow{2}{*}{h-index}} & {\multirow{2}{*}{i100-index}} \\
    & {multiplicity} & {$\geq$ 2 links} & & & {multiplicity} & {$\geq$ 2 links} \\
    \midrule
    {\cellcolor{lightgray} \textsc{Qwen3-30B-A3B}} \\
    \textsc{Wikontic}
        & 1.05 &  4.3 & 339.7 \pm 0.6 & 895.3 \pm 10.8 
        & 1.11 &  8.6 & 341.7 \pm 3.5 & 977.0 \pm  3.6 \\
    \OurInContext
        & {\CircledText{1.18}} & {\CircledText{14.1}} & \bfseries 515.7 \pm 5.0 & \bfseries 1821.0 \pm 19.1
        & {\CircledText{1.26}}\ & {\CircledText{17.6}}\ \ \ & \bfseries 643.0 \pm 2.6 & \bfseries 2491.7 \pm 35.7  \\
    \OurMethod
        & 1.04 &  3.2 & 336.0 \pm 1.0 &  884.7 \pm  4.2
        & 1.10 &  7.8 & 364.0 \pm 1.7 & 1031.7 \pm  6.8 \\
    \OurMethodCorrect
        & 1.03 &  2.7 & {\ulnum{402.7 \pm 2.5}} & {\ulnum{1141.7 \pm 16.8}}
        & 1.08 &  6.6 & {\ulnum{430.3 \pm 2.5}} & {\ulnum{1347.0 \pm  4.6}}  \\
    \midrule
    {\cellcolor{lightgray} \textsc{GPT-OSS-120B}} \\
    \textsc{Wikontic}
        & 1.05 &  4.4 & {\ulnum{370.3 \pm 3.1}} & \bfseries 1068.0 \pm 8.5
        & 1.11 &  9.0 & {\ulnum{401.7 \pm 3.1}} & \bfseries 1182.0 \pm 3.0 \\
    \OurInContext
        & 1.02 &  2.0 & 320.7 \pm 2.1 & 775.3 \pm 8.0
        & 1.05 &  4.7 & 337.3 \pm 6.7 & 902.0 \pm 7.9 \\
    \OurMethod
        & 1.04 &  3.2 & 329.0 \pm 3.5 &  823.3 \pm 16.5
        & 1.13 & 10.1 & 344.3 \pm 1.2 &  901.7 \pm  5.8 \\
    \OurMethodCorrect
        & 1.03 &  2.6 & \bfseries 383.3 \pm 6.7 & \bfseries 1054.7 \pm 25.5
        & 1.10 &  8.1 & \bfseries 408.7 \pm 3.5 & {\ulnum{1157.0 \pm 9.2}} \\
    \bottomrule
    \end{tabular}
\caption{\textbf{\OurMethodCorrect extracts KGs containing graph patterns useful to construct and run SPARQL queries.} For each method we show our h-index and i100-index metrics computed on the patterns frequencies, as well as edge multiplicity metrics measuring the ambiguity in choosing the right predicate between two entities (see \cref{sec:emprical-evaluations-patterns}).
On \textsc{Qwen3-30B-A3B}, the \textsc{In-Context} method achieves an outlier edge multiplicity metrics (see circled numbers), thus boosting the number of pattern matches by extracting redundant-triples (see also \cref{tab:qualitative-redundant-triples}).
\revision{%
\cref{tab:patterns-hotpot-i1-i5-index} complements these results with i1-index and i5-index metrics, which reward BGP diversity more when compared to h-index and i100-index, showing \OurMethodCorrect being competitive in terms of graph patterns diversity.
}%
}%
    \label{tab:empirical-evaluation-patterns}
\end{table*}

\subsection{Q2. Question Answering Experiments}\label{sec:empirical-evaluation-qa}

We evaluate the capability of our method in encoding knowledge useful for QA.
For all methods we adopt the multi-step 
QA
approach designed in \citet{chepurova2026wikontic}, which iteratively decomposes a multi-hop question into simpler 1-hop ones each being answered with an LLM.
When answering to a 1-hop question, the LLM is asked to extract relevant entities from the question, which are then linked to entities in the extracted KG via embedding similarity.
The linked entities are used to retrieve the associated triples and qualifiers, which provide the context for answering the question.
Importantly, for each question, we do not keep separate KGs for its supporting texts; instead, we merge the knowledge extracted from all supporting texts into a single unified KG before performing QA.

Following the literature on QA, in \cref{tab:empirical-evaluation-qa} we report exact match (EM) and $\text{F}_1$ metrics computed between
predicted and ground-truth answers, excluding articles.
\OurMethod achieves the best or second-best QA results, thus being competitive with \textsc{KGGen}, \textsc{Wikontic} and the \OurInContext baseline.
Moreover, applying ontology-based corrections causes little or no loss in QA performance, suggesting that the corrections preserve the semantics of the extracted triples and qualifiers.
\revision{%
In \cref{tab:empirical-evaluation-qa} we also report QA performances achieved by text-based methods.
That is, vector RAG and HippoRAGv2 answer questions by directly retrieving and reasoning over the original source text rather than operating exclusively on symbolic triples and their qualifiers like \OurMethod.
While we observe that operating directly on the text can boost QA performances on some configurations (e.g., vector RAG with \textsc{GPT-OSS-120B} on HotpotQA), these methods do not support symbolic reasoning.
}%
Finally, \cref{tab:empirical-evaluation-qa-no-qualifiers} shows an ablation on the qualifiers, further supporting the finding of \citet{chepurova2026wikontic} that qualifiers boost QA performances.

\subsection{Q3. Patterns Enabling SPARQL Queries}\label{sec:emprical-evaluations-patterns}

We evaluate how amenable the extracted KGs are to SPARQL queries, for which consistency with the underlying ontology is crucial (see \cref{sec:background-ontology-constraints}).
To disentangle our evaluation with the challenging task of translating questions in natural language to SPARQL
\citep{gashkov2025sparql,perevalov2025text-to-sparql}, we introduce a benchmark measuring the occurrence of \emph{basic graph patterns} (BGPs). BGPs are abstract sub-graphs defined as a conjunction of triples forming the \verb|WHERE| clause in a SPARQL query (e.g., the query in \cref{sec:background-ontology-constraints}).
Hence, their presence in the extracted KGs is a necessary condition for executing SPARQL queries.%

\textbf{BGPs datasets.}
We build a dataset of BGPs of queries executed on the Wikidata endpoint, which we extract from the LSQ-2.0 collection \citep{saleem2015lsq,stadler2024lsq2}.
This yields a set of a few thousands BGPs over hundreds of predicates.
Moreover, to obtain a larger set of patterns, we artificially generate BGPs from the ontology domain-range constraints using templates common in the complex query answering literature \citep{gregucci2025complex}.
We detail the procedures and the statistics of the two BGPs datasets in \cref{app:patterns-datasets}.

\textbf{Metrics.}
One way to evaluate the extracted KGs would be to simply compare the total frequency of BGP matches.
However, since some BGPs appear much more frequently than others, we use metrics that also account for \emph{rare} BGPs.
Inspired by the h-index and i10-index metrics, which reward a consistent output of highly cited papers, we adopt analogous metrics to reward methods that match many BGPs with consistently high frequency across a \textit{diverse set of} patterns.
In \cref{tab:empirical-evaluation-patterns} we report h-index and i100-index metrics computed on all patterns, together with the average number of links connecting two entities (edge multiplicity), and the the percentage of connected subject-object pairs with two or more links.
The last two metrics serve as a sanity check for redundant triples that encode the same information with different predicates, since such redundancy could artificially boost BGPs matches at the cost of introducing ambiguity.

\textbf{Results.}
Our \OurMethodCorrect on \textsc{Qwen3-30B-A3B} is second only to the \OurInContext baseline, which however has high edge multiplicity.
As we detail in \cref{tab:qualitative-redundant-triples}, this is because \OurInContext extracts highly-redundant triples over pairs of entities encoding the same information, yet with different predicates that have slightly different meaning.
Instead, on \textsc{GPT-OSS-120B}, our \OurMethodCorrect achieves similar performances to \textsc{Wikontic}, and much better than \OurInContext, without having an outlier edge multiplicity.
In all cases our ontology-based corrections boost h-index.
In \cref{tab:breakdown-patterns-hotpot} and \cref{tab:breakdown-patterns-musique} we report the results achieved on only the LSQ-2.0-based or artificial patterns, respectively.

\section{Conclusion}\label{sec:conclusion}

We introduced a method for ontology-grounded KG extraction that corrects ontology violations \emph{ex-post}, improving token efficiency compared to existing methods based on repeated LLM calls or long ontology-aware contexts.
Our approach is supported by experiments evaluating the many facets of KG extraction: ontology consistency, QA performances, token efficiency, and suitability for symbolic querying.
In particular, we have shown that targeted LLM-predicted corrections substantially improve ontology consistency while preserving semantic accuracy. 
To the best of our knowledge, this is the first work to study KG extraction through such a broad empirical evaluation, while also providing an in-depth evaluation over the Wikidata ontology.
Finally, we believe our SPARQL graph-patterns benchmark 
can serve as a useful tool for quantitatively assessing how well a KG extraction method supports SPARQL-based
querying.

\section*{Limitations}

Our experiments focused on the general-purpose ontology available in Wikidata, as this is a popular setting in the KG extraction community.
Instead, one can consider custom-build ontologies that may be specific for a certain domain, e.g., for biomedical KGs \citep{walsh2020biokg}.
We believe understanding whether the performances achieved by LLMs generalize to custom or even closed-source ontologies remains poorly understood.
In addition, our method requires a detailed ontology specification, including both domain-range and qualifiers constraints, which might not be available.

Moreover, in our experiments we restricted ourselves to open-weight LLMs only.
While this promotes transparency and reproducibility over time, further experiments exploiting closed-weight yet more powerful LLMs could provide a more detailed overview of the performances achieved by the different KG extraction methods.

\revision{%
The lack of an ontology-aligned ground-truth KG for the considered benchmarks does not allow us to evaluate our simplifying assumptions for the efficient canonicalization of types and predicates (\cref{sec:canonicalization-step}).
Nevertheless, our qualitative analysis in \cref{app:human-qualitative-evaluation} partially addresses this by showing that the canonicalized entity types are faithful w.r.t. the source text in 80-85\% of the analyzed triples.
}%

Lastly, our experiments do not show QA performances achieved by means of SPARQL queries executed on the extracted KGs.
This is because it would require leveraging text-to-SPARQL translation techniques to convert questions in natural language to SPARQL queries \citep{gashkov2025sparql,perevalov2025text-to-sparql}.
Since this is already a very challenging open problem, in our work we instead rely on proxy metrics based on the occurrences of SPARQL query graph patterns alone.

\revision{%
\section*{Acknowledgements}
We would like to acknowledge Antonio Vergari for meaningful discussions during the early stages of this work.
We also thank the anonymous reviewers for their constructive feedback during the rebuttal.
}%

\bibliography{./referomnia}

\cleardoublepage
\appendix

\counterwithin{table}{section}
\counterwithin{figure}{section}
\counterwithin{algorithm}{section}
\renewcommand{\thetable}{\thesection.\arabic{table}}
\renewcommand{\thefigure}{\thesection.\arabic{figure}}
\renewcommand{\thealgorithm}{\thesection.\arabic{algorithm}}

\section{Datasets and Ontology}\label{app:datasets-ontology}

\textbf{Question answering datasets.}
We use the same dataset splits of HotpotQA \citep{yang2018hotpotqa} and MuSiQue \citep{trivedi2022musique} used in \citet{gutierrez2025from,chepurova2026wikontic}.
Each split is a subset of the original dataset consisting of 1000 entries each.
Each dataset entry contains a list of text paragraphs as well as a question expressed in natural language.
All KG extraction methods are run on each text paragraph, and the multiple KGs obtained for each entry are merged by also leveraging the respective entity deduplication steps.
Finally, before running our QA experiments, the KGs extracted from each entry are merged.

\textbf{Ontology.}
We extract a subset of entity types, predicates, and ontology constraints from Wikidata \citep{vrandeco2014wikidata}.
We do so by re-adapting the code by \citet{chepurova2026wikontic} using SPARQL queries run on the \href{https://query.wikidata.org/}{public Wikidata endpoint}.
That is, we also extract constraints defined over qualifiers (see \cref{sec:background-ontology-constraints}).
That is, for each predicate $r$, we also query for the set of allowed qualifier predicates, if any.
In \cref{tab:ontology-statistics} we show statistics about the retrieved ontology, such as the number of types, predicates, and constraints.
If a predicate does not specify constraints, we assume it has none.
Finally, we ensure we use the same ontology fragment for the baselines \textsc{Wikontic} and \textsc{In-Context} considered in \cref{sec:empirical-evaluation}.

\begin{table}[h]
\centering
\setlength{\tabcolsep}{5.5pt}%
\tablefont%
\begin{tabular}{ccccccc}
\toprule
\multicolumn{3}{c}{\textbf{Ontology Elements}}
    & \multicolumn{3}{c}{\textbf{Ontology Constraints for Predicates}} \\
\multirow{2}{*}{$|\calT|$}
    & \multirow{2}{*}{$|\calR|$}
    & \multirow{2}{*}{$|\calQ|$}
    & \multirow{2}{*}{\textbf{w/ Domain}}
    & \multirow{2}{*}{\textbf{w/ Range}}
    & \textbf{w/ Allowed} \\
    & & & & & \textbf{Qualifiers} \\
\midrule
3768 & 2700 & 786 & 1757 & 1926 & 496 \\
\bottomrule
\end{tabular}%
\caption{\textbf{Statistics of the Wikidata ontology fragment used.} We show the number of entity types $|\calT|$, the total number of predicates $|\calR$|, and the number of qualifier predicates $\calQ\subset\calR$, i.e., $|\calQ|$. In addition, we show how many predicates in $\calR$ specify domain, range, and allowed qualifier predicates ontology constraints.}
\label{tab:ontology-statistics}
\end{table}

\begin{figure*}[h]
    \centering%
    \includegraphics[width=0.99\linewidth]{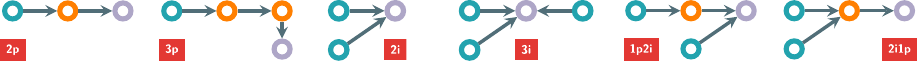}
    \caption{\textbf{Examples of basic graph patterns (BGPs) structures associated to different queries} (the image is taken from \citet{gregucci2025complex}). Each sub-graph structure corresponds to a particular type of multi-hop query.
    For instance, a query asking for ``Which actor performed in a movie filmed in New York City'' is a 2-hop SPARQL query (or \textsf{2p}) having the following BGP with ground entities replaced with variables: $\{ \ {\color{petroil2} ?x1} \ \textsf{cast member:} \ {\color{gold4} ?x2} \ .\ \  {\color{gold4} ?x2} \ \textsf{filming location:} \ {\color{lacamlilac} ?x3} \ .\ \}$, where \textsf{cast member:} (P161) and \textsf{filming location:} (P915) are two Wikidata predicates \citep{vrandeco2014wikidata}.
    We artificially construct BGPs having structures such as the ones above.
    As detailed in \cref{app:patterns-datasets}, we do so by extracting such patterns from a special KG linking entity types in the ontology by means of domain-range constraints.
    We also consider alternative BGPs structures where the directionality of predicates is inverted (e.g., yielding the \textsf{r-2i} structure from \textsf{2i}).
    \cref{tab:artificial-query-pattern-descriptions} shows a formalization of each BGP structure in SPARQL, as well as their number.}
    \label{fig:query-patterns}
\end{figure*}

\section{Basic Graph Patterns Benchmark}\label{app:patterns-datasets}

In order to evaluate the quality of the KGs extracted by ontology-grounded methods, we compute the occurrence of \emph{basic graph patterns} (BGPs) useful to build SPARQL queries.
For example, a BGP consisting of two abstract triples and using predicates \verb|occupation:| and \verb|awardReceived:| is:
\begin{minted}[autogobble,framesep=0pt,fontsize=\small]{sparql}
    { ?person occupation: ?occupation .
      ?person awardReceived: ?award . }
\end{minted}
This abstract BGP corresponds to a sub-graph pattern (or graph motif) with three entity variables (\verb|?person|, \verb|?occupation|, \verb|?award|) and two predicates (\verb|occupation:| and \verb|awardReceived:|), and can be the skeleton of a \verb|WHERE| statement in a SPARQL query.
Alternatively, following the literature of query answering on incomplete KGs \citep{gregucci2025complex}, the same BGP can be rewritten as the existentially-qualified first-order formula:
\begin{align*}
    \exists P,O,A\colon & \textsf{occupation}(P,O)\ \land \\
    & \textsf{awardReceived}(P,A).
\end{align*}
The idea of our benchmark based on BGPs is to measure how many times this graph pattern matches with an extracted KG, i.e., counting the number of distinct assignments to the person, occupation and award entity variables.
The higher is the number of matching BGPs, the higher is the likelihood that a SPARQL query having such BGP will return a non-empty result.
In other words, the existence of BGPs in a KG is a necessary---although not sufficient---condition for retrieving relevant information by means of a SPARQL query.

This setting is similar to another one used to evaluate graph generative models, e.g., where the percentage of graphs having certain graph motifs as in molecules generation is computed \citep{li2018learning-graphs}.
Similarly, other works compare the divergence between the distribution of the number of graph motifs in the ground truth and in the generated graphs \citep{demsar2014combinatorial,you2018graphrnn}.
However, the difference is that in our case we do not have access to the ground truth KG.

Therefore, inspired by h-index and i$k$-index citation metrics rewarding an author for having many citations over a diverse set of papers, we use the same metrics to reward a KG extraction method for having a high number of matches over a diverse set of BGPs.
In our case, a method achieves an h-index of $n$ if there are at least $n$ BGPs appearing with a frequency of $n$ or above in the extracted KGs.
Moreover, a method achieves an i$k$-index of $n$ if there are at least $n$ BGPs with a frequency equal or greater than $k$.
To compute the number of matching BGPs, we filter out the extracted triples violating the ontology, and recover the frequency of each BGP using a SPARQL run on a \href{https://github.com/ad-freiburg/qlever}{qlever SPARQL endpoint}.
Next, we detail two datasets of BGPs we use: one extracted from real-world queries run on the Wikidata SPARQL endpoint, and one artificially constructed from the considered domain-range ontology constraints.
\cref{tab:query-patterns-stats} shows statistics about both BGPs datasets.

\begin{table}[!t]
\tablefont%
\centering%
\begin{tabular}{llrr}
\toprule
\textbf{BGPs Dataset} & & \textbf{Artificial} & \textbf{LSQ-2.0} \\
\midrule
\multicolumn{2}{l}{\textbf{\# patterns}}              & 264259 &  5757 \\
\multicolumn{2}{l}{\textbf{\# predicates}} &    266 &   716 \\
\midrule
\multirow{3}{*}{\textbf{\# Triples}}
    & Min. &  2 &  2 \\
  & Avg. &  2.946 &  2.905 \\
  & Max. &  3 & 14 \\
\midrule
\multirow{3}{*}{\textbf{\shortstack[l]{Pattern degree\\ (or \# Joins)}}}
    & Min. &  2 &  2 \\
  & Avg. &  2.000 &  2.724 \\
  & Max. &  2 & 14 \\
\bottomrule
\end{tabular}
\caption{\textbf{Statistics of basic graph patterns (BGPs) from our artificially-constructed dataset and from the real-world Wikidata queries from LSQ-2.0.} We show the number of BGPs and the number of unique predicates. Moreover, we show the minimum/average/maximum number of triples per BPG, as well as the number of joins (i.e., the node degree of the BGP itself). The BGPs obtained from the LSQ-2.0 collection \citep{saleem2015lsq,stadler2024lsq2} consists of a more diverse set of predicates when compared to the artificially extracted BPGs (see \cref{app:patterns-datasets}).}
    \label{tab:query-patterns-stats}
\end{table}

\begin{table}[!t]
    \centering%
    \tablefont%
    \begin{tabular}{llr}
    \toprule
    \textbf{Kind} & \textbf{Basic Graph Pattern Formalization} & \textbf{Count} \\
    \midrule
    2p   & {\color{petroil2} ?x1} ?r1 {\color{gold4} ?x2} . {\color{gold4} ?x2} ?r2 {\color{lacamlilac} ?x3} . & 6'427 \\
    2i   & {\color{petroil2} ?x1} ?r1 {\color{lacamlilac} ?x3} . {\color{petroil2} ?x2} ?r2 {\color{lacamlilac} ?x3} . & 3'041 \\
    1p2i & {\color{petroil2} ?x1} ?r1 {\color{gold4} ?x2} . {\color{gold4} ?x2} ?r2 {\color{lacamlilac} ?x4} . {\color{petroil2} ?x3} ?r3 {\color{lacamlilac} ?x4} . & 10'000 \\
    2i1p & {\color{petroil2} ?x1} ?r1 {\color{gold4} ?x3} . {\color{petroil2} ?x2} ?r2 {\color{gold4} ?x3} . {\color{gold4} ?x3} ?r3 {\color{lacamlilac} ?x4} . & 10'000 \\
    3p & {\color{petroil2} ?x1} ?r1 {\color{gold4} ?x2} . {\color{gold4} ?x2} ?r2 {\color{gold4} ?x3} . {\color{gold4} ?x3} ?r3 {\color{lacamlilac} ?x4} . & 10'000 \\
    3i & {\color{petroil2} ?x1} ?r1 {\color{lacamlilac} ?x4} . {\color{petroil2} ?x2} ?r2 {\color{lacamlilac} ?x4} . {\color{petroil2} ?x3} ?r3 {\color{lacamlilac} ?x4} . & 10'000 \\
    r-2i   & {\color{petroil2} ?x1} ?r1 {\color{lacamlilac} ?x2} . {\color{petroil2} ?x1} ?r2 {\color{lacamlilac} ?x3} . & 4'791 \\
    r-1p2i & {\color{petroil2} ?x1} ?r1 {\color{gold4} ?x2} . {\color{petroil2} ?x1} ?r2 {\color{lacamlilac} ?x3} . {\color{gold4} ?x2} ?r3 {\color{lacamlilac} ?x4} . & 10'000 \\
    r-2i1p & {\color{petroil2} ?x1} ?r1 {\color{gold4} ?x2} . {\color{gold4} ?x2} ?r2 {\color{lacamlilac} ?x3} . {\color{gold4} ?x2} ?r3 {\color{lacamlilac} ?x4} . & 
    10'000 \\
    r-3i & {\color{petroil2} ?x1} ?r1 {\color{lacamlilac} ?x2} . {\color{petroil2} ?x1} ?r2 {\color{lacamlilac} ?x3} . {\color{petroil2} ?x1} ?r3 {\color{lacamlilac} ?x4} . & 10'000 \\
    \bottomrule
    \end{tabular}
    \caption{\textbf{Artificial BGPs structures and their SPARQL formalization.} We extract a number of BGPs for each structure starting from an ad-hoc graph consisting of entity types as nodes where links encodew domain-range constraints (see \cref{app:patterns-datasets}).
    For computational reasons, we limit the number of BGPs to $10^4$.}
    \label{tab:artificial-query-pattern-descriptions}
\end{table}

\textbf{Real-world BGPs.}
We extract BGPs from real-world SPARQL queries on Wikidata, which are collected under the Linked Dataset SPARQL Query Logs (LSQ-2.0) \citep{saleem2015lsq,stadler2024lsq2}.
We do so by running SPARQL queries on the \href{https://lsq.data.dice-research.org/sparql}{LSQ-2.0 publicly available endpoint}.
After we collect a BGP, we replace constant entities in it with fresh variables and filter it out only if it contains multiple triples with the same predicate.
This allows us to only keep interesting BGPs that do not map to a specific SPARQL query with bound entities.
We then remove duplicate BGPs via a directed graph isometry test.
The obtained dataset consists of 5757 BGPs over 716 Wikidata predicates.

\textbf{Artificial BGPs.}
To build a larger dataset of BGPs, we artificially construct them by exploiting common Wikidata predicates and BGP structures commonly used in existing complex query answering benchmarks \citep{gregucci2025complex}.
We start by building a special KG---called \emph{ontology KG}---encoding the domain-range constraints over a set of popular Wikidata predicates, which we take from the DocRED \citep{yao2019docred} and CodRED \citep{yao2021codred} datasets.
That is, we construct a KG such that for each predicate $r$ specifying at least one between the domain and range constraints, we insert a triple $(t_s,r,t_o)$ where $t_s\in\domain(r)$ (resp. $t_o\in\range(r)$).
If the domain (resp. range) is \emph{not} specified, we introduce a fresh \emph{``any''} type node in the ontology KG.
Then, we run SPARQL queries on such KG in order to exhaustively recover BGPs with ground predicates having various structures corresponding to complex queries.
\cref{fig:query-patterns} shows the different BGP structures we consider useful to construct SPARQL queries.
For computational reasons we randomly sample up to $10'000$ BGPs per structure.
This yields a set of BGPs having various kinds of structures and with different combinations of predicates that are consistent withe domain-range ontology constraints.
We summarize the BGPs structures and their count in \cref{tab:artificial-query-pattern-descriptions}.

\section{Detailed Description of the Qualifiers Correction Step}\label{app:detailed-corrections}

After we corrected the triples violating domain-range constraints in \cref{sec:our-method-corrections-step-correct-triples}, we can now correct any attached qualifier violating the ontology.
Similarly to our description of triples correction in \cref{sec:our-method-corrections-step-correct-triples}, here we describe a collection of possible KG extraction mistakes that make qualifiers violating the ontology.
Each of these possible mistakes will then motivate the corrective actions that an LLM will be prompted to choose from.

\noindent%
\textbf{V4.}
A qualifier $(r_q,o_q)$ that is attached to a triple $(s,r,o)$ is inconsistent because $r_q$ is \textbf{not a qualifier predicate allowed by the triple predicate} $r$, i.e., $r_q\not\in\qualifiers(r)$.
For example, in Wikidata we have that $\textsf{location}\notin\qualifiers(\textsf{publication date})$, i.e., \textsf{location} is not a qualifier predicate allowed by \textsf{publication date}.
Therefore, the qualifier $(\textsf{location}, \textsf{London})$ cannot be a consistent when attached to the triple $(\textsf{Animal Farm},\textsf{publication date},\textsf{1945})$.
Instead, $\textsf{place of publication}\in\qualifiers(\textsf{publication date})$ is an allowed qualifier and therefore it should be used instead of $\textsf{location}$ as qualifier predicate.
The corrective action here is \emph{replacing the qualifier predicate with one allowed by the predicate in the triple}.

\noindent%
\textbf{V5.}
A qualifier $(r_q,o_q)$ of a triple $(s,r,o)$ might not be consistent with respect to the range constraint of $r_q$, and the reason is that $r_q$ is \textbf{a qualifier predicate that has been mis-canonicalized}.
That is, we have that $r_q\in\qualifiers(r)$, yet the types of the entity $o_q$ are incosnsistent w.r.t. the range of $r$.
For example, consider the qualifier $(\textsf{distribution format}, \textsf{Penguin Books})$ attached $(\textsf{Animal Farm},\textsf{publication date},\textsf{1945})$.
This qualifier is inconsistent if $\tau(\textsf{Penguin Books}) = \{\textsf{publisher}\}$, since it would not satisfy the range constraint of \textsf{distribution format}, i.e., expecting a format.
Yet, we have that $\textsf{distribution format}\in\qualifiers(\textsf{publication date})$.
The correcting action is replacing the qualifier predicate with \textsf{publisher}, which is allowed by the predicate \textsf{publication date} and it is consistent with the type of \textsf{Penguin Books}.

\noindent%
\textbf{V6.} Finally, similarly to the underspecified or missing entity types mistake discussed for triples in \cref{sec:our-method-corrections-step-correct-triples}, a qualifier $(r_q,o_q)$ might not be consistent because of the same reason, i.e., the types of $o_q$ are not specific enough, or a type is missing, in order to be consistent with the range of $r_q$.

As detailed in \cref{sec:our-method-corrections-step-correct-qualifiers}, these possible mistakes suggest us to prompt an LLM to select between two syntactic transformations of the qualifier, namely \textsf{replace\_predicate} and \textsf{add\_object\_type}.

\section{LLM-based Baseline with In-Context Ontology Constraints}\label{app:in-context-baseline}

We detail our KG extraction baseline inserting the ontology constraints as part of the input context, which we denote simply as \OurInContext, and report the prompts in \cref{app:prompts-in-context-baseline}.

The idea of \OurInContext is to (1) firstly predict a relevant fragment of the ontology consisting of predicates and allowed qualifiers given the source text; and (2) extract the KG from the source text by instructing the LLM to comply with the ontology.
For (1), we prompt an LLM to extract open-domain predicates given the source text.
Then, the extracted open-domain predicates are mapped to the top-$k$ most similar predicates allowed by ontology, in terms of embedding similarity computed over predicate labels and aliases.
We use $k=20$ in our experiments.
Then, for (2) we insert a verbalization of the predicted predicates, relative domain-range constraints, and allowed qualifiers as part of the LLM context upon KG extraction.
If an extracted predicate (resp. entity type) is not one of the allowed ones (resp. one appearing in a domain or range of a predicate), we still keep it as an \emph{unknown} predicate (resp. entity type) labeled with the same name.

\section{Human Qualitative Evaluation}\label{app:human-qualitative-evaluation}

\revision{%
We perform a qualitative analysis using two of the authors of this paper to understand whether \OurMethodCorrect fixes the ontology violations by sacrificing the semantic faithfulness of triples w.r.t. the source text.
To do so, we focus on HotpotQA and give each human evaluator 100 triples extracted and corrected using \textsc{Qwen3-30B-A3B}, and 100 triples extracted and corrected using \textsc{GPT-OSS-120B}.
The qualitative evaluation is therefore performed by considering a total of 400 triples that have been randomly sampled from the KGs.

\textbf{Evaluation format.}
A human evaluator receives the following information about a triple correction:
\begin{itemize}[itemsep=0pt]
    \item The triple violating the ontology constraints, i.e., extracted by \OurMethod and \emph{before} the correction with MEND.
    \item The types of subject and object entities \emph{before} the correction.
    \item The triple satisfying the ontology constraints, i.e., \emph{after} the correction with MEND.
    \item The types of subject and object entities \emph{after} the correction.
    \item The source text fragment where the triple was originally extracted from.
\end{itemize}
Then, we ask the evaluator to answer either \emph{Yes} or \emph{No} to the following questions while considering the given source text fragment:
\begin{itemize}[itemsep=0pt]
    \item Is the inconsistent triple \emph{before} the correction faithful?
    \item Are the subject and object entity types \emph{before} the correction faithful?
    \item Is the consistent triple \emph{after} the correction faithful?
    \item Are the subject and object entity types \emph{after} the correction faithful?
\end{itemize}
In other words, each triple correction is checked against the source text for semantic faithfulness by one annotator using binary metrics.

\textbf{Results.}
We then compute the percentage of triples and the related entity types that are regarded as faithful \emph{before} and \emph{after} the predicted correction.
Although brief, our qualitative results reported in \cref{tab:human-qualitative-evaluation} suggest that the rate of text-based faithfulness of triples increases after the corrections.
Meanwhile, the faithfulness of the subject and object entity types remains approximately the same.
These results focusing on semantic faithfulness complement our experiments on downstream QA (see \cref{sec:empirical-evaluation-qa}), where we quantitatively show that applying post-extraction corrections comes with little to no performances loss.
}%

\begin{table}[!t]
\tablefont%
\setlength{\tabcolsep}{2pt}%
\begin{tabular}{lcc}
\toprule
& \textsc{Qwen3-30B-A3B} & \textsc{GPT-OSS-120B} \\
\midrule
\textbf{Before the correction} & & \\
\% faithful triples            & 37.0 & 40.5 \\
\% triples w/ faithful types   & 86.0 & 79.0 \\
\textbf{After the correction}  & & \\
\% faithful triples            & 65.5 & 67.0 \\
\% triples w/ faithful types   & 85.5 & 80.0 \\
\bottomrule
\end{tabular}
\caption{%
\revision{%
\textbf{Our framework based on post-extraction corrections of ontology violations does not sacrifice semantic faithfulness.}
On HotpotQA, we show the percentage of triples and subject and object entity types that are regarded as faithful w.r.t. the source text in our human audit \emph{before} and \emph{after} the corrections (see \cref{sec:our-method-corrections-step-correct-triples}).
On both LLMs that we considered in our experiments, the percentage of faithful triples increases after the corrections, while the percentage of triples whose both subject and object entity types are text-aligned remains approximately the same.
}%
}%
\label{tab:human-qualitative-evaluation}
\end{table}

\section{Additional Experimental Results}\label{app:additional-results}

\cref{tab:empirical-evaluation-kgc-stats-musique} shows KG extraction statistics, ontology consistency metrics and token cost achieved on MuSiQue instead.
The observations made for HotpotQA in \cref{sec:empirical-evaluation-consistency}, i.e., competitive ontology consistency rates with a modest token cost, translate to MuSiQue as well.
\cref{tab:empirical-evaluation-qa-no-qualifiers} reports QA results ablated by the qualifiers presence, confirming their importance in capturing crucial semantic information from the text.
In \cref{tab:breakdown-patterns-hotpot,tab:breakdown-patterns-musique} we separate the results obtained on LSQ-2.0-based and artificial BGP datasets (see \cref{app:patterns-datasets}).
Finally, in \cref{tab:qualitative-redundant-triples} we provide a qualtiative analysis of triples extracted by the \OurInContext baseline on \textsc{Qwen3-30B-A3B}, confirming our claim in \cref{sec:emprical-evaluations-patterns} that triples containing redundant information are extracted, thus boosting h-index metrics on our BGPs benchmark.

\textbf{Hardware, time, LLMs and repetitions.}
All experiments are executed on NVIDIA RTX A6000 and A100 GPUs, and results can be reproduced in a about a week.
We use the default sampling hyperparameters for all LLMs.
We do so by using vLLM to serve our models.
All the experimental results we show in this paper are obtained by averaging across 3 independent runs.

\clearpage%

\begin{table*}[!t]
    \centering%
    \tablefont%
    \setlength{\tabcolsep}{6pt}%
    \begin{tabular}{%
    l
    S[table-format=6(4), separate-uncertainty=true]
    S[table-format=2.1(1.1), separate-uncertainty=true]
    S[table-format=5(3), separate-uncertainty=true]
    S[table-format=2.1(1.1), separate-uncertainty=true]
    S[table-format=3.1]
    S[table-format=2.1]
    S[table-format=2.1]
    @{~(} 
    S[table-format=1.2]
    @{)}
    }
    \toprule
    \textbf{Dataset: MuSiQue} &
    \multicolumn{2}{c}{\textbf{Triples Statistics}} &
    \multicolumn{2}{c}{\textbf{Qualifiers Statistics}} &
    \multicolumn{3}{c}{\textbf{\# Tokens $(\times 10^6)$}} \\
    & {\# Total} & {\% Valid} & {\# Total} & {\% Valid} & {Prompt} & {Completion} & \multicolumn{2}{c}{Weighted} \\
    \midrule
    {\cellcolor{lightgray} \textsc{Qwen3-30B-A3B}} \\
    \textsc{KGGen}
        & 283729 \pm  992 &        {---} &     0 \pm    0 &        {---} & 140.6 &  20.7 &  55.8 & 0.66 \\
    \textsc{Wikontic}
        & 204689 \pm  207 & 72.9 \pm 0.2 & 88209 \pm  540 &        {---} & 260.8 &  18.9 &  84.1 & 1.00 \\
    \OurInContext
        & 345328 \pm 1621 & {\ulnum{78.2 \pm 0.2}} & 75205 \pm  505 & {\ulnum{83.3 \pm 0.1}} & 176.4 &  25.5 &  69.6 & 0.83 \\
    \OurMethod
        & 224294 \pm  619 & 58.7 \pm 0.2 & 89503 \pm 1563 & 29.3 \pm 0.1 &  65.5 &  14.0 &  \bfseries 30.4 & 0.36 \\
    \OurMethodCorrect
        & 222749 \pm  636 & \bfseries 95.7 \pm 0.1 & 89503 \pm 1563 & \bfseries 90.1 \pm 0.2 &  80.0 &  12.5 & \ \ {\ulnum{32.5}} & 0.39 \\
    \midrule
    {\cellcolor{lightgray} \textsc{GPT-OSS-120B}} \\
    \textsc{KGGen}
        & 230359 \pm 6556 &        {---} &     0 \pm   0 &        {---} &  166.3 & 103.7 & 145.3 & 0.83 \\
    \textsc{Wikontic}
        & 189396 \pm  517 & \bfseries 99.1 \pm 0.1 & 79261 \pm 375 &        {---} &  269.6 & 101.8 & 176.0 & 1.00 \\
    \OurInContext
        & 112207 \pm  713 & {\ulnum{97.3 \pm 0.1}} & 18610 \pm 407 & \bfseries 98.0 \pm 0.1 &  181.9 &  39.5 & \ \ {\ulnum{85.0}} & 0.48 \\
    \OurMethod
        & 195568 \pm  787 & 63.3 \pm 0.1 & 85701 \pm 484 & 32.2 \pm 0.3 &   66.6 &  54.8 &  \bfseries 71.4 & 0.41 \\
    \OurMethodCorrect
        & 194539 \pm  800 & {\ulnum{97.3 \pm 1.0}} & 85701 \pm 484 & {\ulnum{94.8 \pm 0.1}} &  135.4 & 102.6 & 136.5 & 0.78 \\
    \bottomrule
    \end{tabular}
\caption{\textbf{\textsc{\OurMethodCorrect} achieves high ontology consistency rates for both triples and qualifiers, yet requiring modest token cost.}
We show the number of triples and qualifiers extracted by different KG construction methods, as well as the percentage consistent with ontology constraints (see \cref{sec:background-ontology-constraints}), where applicable.
We also show the total number of prompt and completion tokens (in millions), and a weighted cost combination that takes into account a prompt-to-completion cost ratio of $\sfrac{1}{4}$ (e.g., as the \$ price ratio for GPT-4.1).}
\label{tab:empirical-evaluation-kgc-stats-musique}
\end{table*}

\begin{table*}[!t]
\begin{minipage}{0.45\linewidth}
    \caption{\textbf{The absence of qualifiers significantly reduces QA performances.} We show the drop of exact match and $\mathrm{F}_1$ metrics for QA achieved by different KG extraction methods and on different datasets, where we remove the triples qualifiers.}
    \label{tab:empirical-evaluation-qa-no-qualifiers}
\end{minipage}%
\hspace{20pt}%
\begin{minipage}{0.5\linewidth}
    \tablefont%
    \setlength{\tabcolsep}{2.5pt}%
    \begin{tabular}{%
    l
    S[table-format=2.1(1.1), separate-uncertainty=true]
    S[table-format=2.1(1.1), separate-uncertainty=true]
    S[table-format=2.1(1.1), separate-uncertainty=true]
    S[table-format=2.1(1.1), separate-uncertainty=true]}
    \toprule
    {\multirow{3}{*}{\textbf{Method}}} &
    \multicolumn{2}{c}{\textbf{HotpotQA}} &
    \multicolumn{2}{c}{\textbf{MuSiQue}} \\
    & \multicolumn{4}{c}{\cellcolor{lightgray} \textsc{GPT-OSS-120B}} \\
    & {{EM}} & {{F\textsubscript{1}}} & {{EM}} & {{F\textsubscript{1}}} \\
    \midrule
    \OurInContext
        & 33.1 \pm 0.7 & 41.8 \pm 0.8 & 17.5 \pm 0.5 & 23.7 \pm 0.7 \\
    \textsc{-- Qualifiers}
        & 32.8 \pm 0.5 & 40.2 \pm 0.4 & 16.1 \pm 1.0 & 21.9 \pm 1.3 \\
    \midrule
    \textsc{Wikontic}
        & {\ulnum{40.4 \pm 0.8}} & {\ulnum{50.4 \pm 0.8}} & {\ulnum{19.9 \pm 0.2}} & {\ulnum{28.0 \pm 0.6}} \\
    \textsc{-- Qualifiers}
        & 35.8 \pm 0.3 & 44.9 \pm 0.1 & 17.2 \pm 0.5 & 23.3 \pm 0.5 \\
    \midrule
    \OurMethodCorrect
        & \bfseries 46.1 \pm 2.0 & \bfseries 56.8 \pm 1.7 & \bfseries 21.5 \pm 0.1 & \bfseries 29.9 \pm 0.2 \\
    \textsc{-- Qualifiers}
        & 38.5 \pm 1.2 & 48.1 \pm 1.4 & 17.4 \pm 0.3 &  23.4 \pm 0.1 \\
    \bottomrule
    \end{tabular}
\end{minipage}
\end{table*}

\begin{table*}[!t]
    \centering%
    \tablefont%
    \setlength{\tabcolsep}{4pt}%
    \begin{tabular}{%
    l
    S[table-format=3.1(1.1), separate-uncertainty=true]
    S[table-format=4.1(2.1), separate-uncertainty=true]
    S[table-format=3.1(2.1), separate-uncertainty=true]
    S[table-format=3.1(1.1), separate-uncertainty=true]
    S[table-format=5.1(3.1), separate-uncertainty=true]
    S[table-format=3.1(2.1), separate-uncertainty=true]
    }
    \toprule
    {\multirow{2}{*}{\textbf{Method}}} &
    \multicolumn{3}{c}{\textbf{LSQ-2.0  Patterns}} &
    \multicolumn{3}{c}{\textbf{Artificial Patterns}} \\
    \cmidrule[0.5pt](lr{0.3em}){2-4} \cmidrule[0.5pt](lr{0.3em}){5-7}
    & {h-index} & {i10-index} & {i100-index} & {h-index} & {i10-index} & {i100-index} \\
    \midrule
    {\cellcolor{lightgray} \textsc{Qwen3-30B-A3B}} \\
    \textsc{Wikontic}
        & 319.7 \pm 0.6 & 1653.3 \pm 17.2 & 699.3 \pm 10.1
        & 132.3 \pm 1.2 & 1542.7 \pm 40.4 & 196.0 \pm 9.6 \\
    \OurInContext
        & \bfseries 469.7 \pm 3.2 & \bfseries 2426.7 \pm 35.5 & \bfseries 1186.7 \pm 3.8
        & \bfseries 237.7 \pm 5.9 & \bfseries 3492.3 \pm 105.3 & \bfseries 634.3 \pm 22.9 \\
    \OurMethod
        & 315.0 \pm 1.0 & 1626.3 \pm 17.5 & 707.7 \pm 1.2
        & 129.0 \pm 1.7 & 1299.7 \pm 19.2 & 177.0 \pm 4.0\\
    \OurMethodCorrect
        & {\ulnum{377.0 \pm 1.7}} & {\ulnum{1957.3 \pm 20.1}} & {\ulnum{866.7 \pm 11.7}}
        & {\ulnum{158.0 \pm 0.1}} & {\ulnum{2204.7 \pm 20.0}} & {\ulnum{275.0 \pm 8.7}} \\
    \midrule
    {\cellcolor{lightgray} \textsc{GPT-OSS-120B}} \\
    \textsc{Wikontic}
        & {\ulnum{346.3 \pm 3.5}} & \bfseries 1884.0 \pm 20.1 & \bfseries 822.3 \pm 9.5
        & \bfseries 148.7 \pm 2.5 & \bfseries 1921.0 \pm 36.5 & \bfseries 245.7 \pm 10.1 \\
    \OurInContext
        & 293.7 \pm 1.5 & 1492.7 \pm 11.0 & 607.7 \pm 2.5
        & 127.7 \pm 2.5 & 1051.7 \pm 18.0 & 167.7 \pm 7.6 \\
    \OurMethod
        & 306.3 \pm 2.5 & 1467.0 \pm 14.0 & 653.3 \pm 18.0
        & 130.3 \pm 1.2 & 1152.0 \pm 41.7 & 169.7 \pm 2.1 \\
    \OurMethodCorrect
        & \bfseries 359.0 \pm 6.1 & {\ulnum{1835.0 \pm 44.8}} & \bfseries 813.0 \pm 22.0
        & \bfseries 150.3 \pm 3.2 & {\ulnum{1863.7 \pm 17.6}} & \bfseries 241.7 \pm 3.5 \\
    \bottomrule
    \end{tabular}
\caption{\textbf{Breakdown of the basic graph pattern (BGPs) matching results on the HotpotQA data set.} We show the total number of pattern matches ($N$), and the h-index, i10-index, and i100-index scores (higher is better).}
\label{tab:breakdown-patterns-hotpot}
\end{table*}

\begin{table*}[!t]
    \centering%
    \tablefont%
    \setlength{\tabcolsep}{4pt}%
    \begin{tabular}{%
    l
    S[table-format=3.1(1.1), separate-uncertainty=true]
    S[table-format=4.1(2.1), separate-uncertainty=true]
    S[table-format=3.1(2.1), separate-uncertainty=true]
    S[table-format=3.1(1.1), separate-uncertainty=true]
    S[table-format=5.1(3.1), separate-uncertainty=true]
    S[table-format=3.1(2.1), separate-uncertainty=true]
    }
    \toprule
        {\multirow{2}{*}{\textbf{Method}}} &
    \multicolumn{3}{c}{\textbf{LSQ-2.0  Patterns}} &
    \multicolumn{3}{c}{\textbf{Artificial Patterns}} \\
    \cmidrule[0.5pt](lr{0.3em}){2-4} \cmidrule[0.5pt](lr{0.3em}){5-7}
    & {h-index} & {i10-index} & {i100-index} & {h-index} & {i10-index} & {i100-index} \\
    \midrule
    {\cellcolor{lightgray} \textsc{Qwen3-30B-A3B}} \\
    \textsc{Wikontic}
        & 325.0 \pm 3.0 & 1778.0 \pm 9.5 & 762.7 \pm 2.9
        & 140.0 \pm 2.0 & 1606.7 \pm 21.1 & 214.3 \pm 4.0  \\
    \OurInContext
        & \bfseries 582.0 \pm 2.6 & \bfseries 2797.0 \pm 13.2 & \bfseries 1519.0 \pm 14.5
        & \bfseries 296.7 \pm 9.0 & \bfseries 4829.7 \pm 64.6 & \bfseries  972.7 \pm 22.2  \\
    \OurMethod
        & 344.3 \pm 0.6 & 1779.7 \pm 19.1 & 797.7 \pm 5.7
        & 146.0 \pm 1.0 & 1459.7 \pm 25.5 & 234.0 \pm 6.1 \\
    \OurMethodCorrect
        & {\ulnum{404.3 \pm 1.5}} & {\ulnum{2153.3 \pm 14.2}} & {\ulnum{994.7 \pm 2.1}}
        & {\ulnum{171.0 \pm 1.0}} & {\ulnum{2640.7 \pm 20.8}} & {\ulnum{352.3 \pm 2.5}} \\
    \midrule
    {\cellcolor{lightgray} \textsc{GPT-OSS-120B}} \\
    \textsc{Wikontic}
        & \bfseries 384.0 \pm 3.6 & \bfseries 2076.7 \pm 29.4 & \bfseries 913.7 \pm 4.2
        & \bfseries 152.7 \pm 4.2 & \bfseries 2205.7 \pm 45.8 & \bfseries 268.3 \pm 1.5 \\
    \OurInContext
        & 314.7 \pm 5.5 & 1689.0 \pm 8.5 & 713.0 \pm 3.0
        & 134.7 \pm 3.1 & 1186.3 \pm 18.9 & 189.0 \pm 5.2   \\
    \OurMethod
        & 324.3 \pm 1.2 & 1620.7 \pm 5.0 & 711.7 \pm 1.5
        & 135.3 \pm 0.6 & 1249.3 \pm 3.2 & 190.0 \pm 4.4 \\
    \OurMethodCorrect
        & \bfseries 385.7 \pm 1.5 & \bfseries 2078.0 \pm 18.5 & {\ulnum{899.7 \pm 9.1}}
        & \bfseries 153.3 \pm 2.1 & {\ulnum{2057.0 \pm 42.4}} & {\ulnum{257.3 \pm 4.9}} \\
    \bottomrule
    \end{tabular}
\caption{\textbf{Breakdown of the basic graph pattern (BGPs) matching results on the MuSiQue data set.} We show the total number of pattern matches ($N$), and the h-index, i10-index, and i100-index scores (higher is better).}
\label{tab:breakdown-patterns-musique}
\end{table*}

\begin{table*}[!t]
\centering%
\tablefont%
\setlength{\tabcolsep}{7pt}%
\begin{tabular}{%
l
S[table-format=5.1(3.1), separate-uncertainty=true]
S[table-format=5.1(3.1), separate-uncertainty=true]
S[table-format=5.1(3.1), separate-uncertainty=true]
S[table-format=5.1(3.1), separate-uncertainty=true]
}
\toprule
\multirow{2}{*}{\textbf{Method}} & \multicolumn{2}{c}{\textbf{HotpotQA}} & \multicolumn{2}{c}{\textbf{MuSiQue}} \\
\cmidrule[0.5pt](lr{0.3em}){2-3} \cmidrule[0.5pt](lr{0.3em}){4-5}
& {i1-index} & {i5-index} & {i1-index} & {i5-index} \\
\midrule
{\cellcolor{lightgray} \textsc{Qwen3-30B-A3B}} \\
\textsc{Wikontic}  &  8383.7 \pm  84.0 & 4436.0 \pm  58.8 & 8246.0 \pm 29.5 & 4603.0 \pm 36.5  \\
\OurInContext      & \bfseries 13353.3 \pm 120.7 & \bfseries 7863.3 \pm 133.5 & \bfseries 15639.7 \pm 125.9 & \bfseries 9837.7 \pm 34.5 \\
\OurMethod         &  7178.0 \pm  37.6 & 3997.7 \pm  52.2 & 7270.7 \pm 43.4 & 4248.7 \pm 9.7  \\
\OurMethodCorrect  & {\ulnum{10378.3 \pm  54.2}} & {\ulnum{5744.3 \pm  43.4}} & {\ulnum{11146.7 \pm 77.7}} & {\ulnum{6434.0 \pm 39.5}}  \\
\midrule
{\cellcolor{lightgray} \textsc{GPT-OSS-120B}} \\
\textsc{Wikontic}  & \bfseries 10037.3 \pm 179.5 & \bfseries 5315.7 \pm 62.1 & \bfseries 11172.3 \pm 222.6 & \bfseries 5957 \pm 76.9   \\
\OurInContext      &  6001.7 \pm  95.3 & 3385.7 \pm 17.9 & 6416.7 \pm 46.7 & 3803.3  \pm 56.1 \\
\OurMethod         &  6587.7 \pm 137.8 & 3573.0 \pm 30.8 & 6728.3 \pm 68.3 & 3783.0  \pm 31.2 \\
\OurMethodCorrect  & \bfseries 9814.0 \pm 140.6 & {\ulnum{5170.3 \pm 42.1}} & {\ulnum{10383.7 \pm 115.8}} & {\ulnum{5633.0 \pm 35.5}} \\
\bottomrule
\end{tabular}
\caption{%
\revision{%
\textbf{Basic graph patterns (BGPs) matching results focusing more on graph patterns diversity.} We complement the results showed in \cref{tab:empirical-evaluation-patterns} with i1-index and i5-index, which reward a wide BGP coverage more when compared to BGP matchings frequency as measured with h-index and i100-index metrics.
For instance, i1-index is the number of BGPs in both LSQ-2.0 and artificial patterns datasets (see \cref{app:patterns-datasets}) that have at least one occurrence in the KGs extracted by the different methods.
Therefore, the higher the i1-index is, the better is the coverage of different BGPs by a KG construction method.
}%
}%
\label{tab:patterns-hotpot-i1-i5-index}
\end{table*}

\begin{table*}[!t]
\centering%
\tablefont%
\setlength{\tabcolsep}{6.5pt}%
\fontsize{8}{10}\selectfont
\renewcommand{\textsf}[1]{\oldtextsf{\fontsize{7}{8}\selectfont #1}}
\begin{tabular}{llll}
\toprule
\textbf{Common Jointly-} & \multirow{2}{*}{\textbf{Count}} & \multirow{2}{*}{\textbf{Example Source Text}} & \multirow{2}{*}{\textbf{Example of Extracted Triples}} \\
\textbf{-extracted Predicates} \\
\midrule
\makecell[l]{%
    \textsf{location of creation (P1071)}\\
    \textsf{country of origin (P495)}\\
} & 598 &
\makecell[l]{%
    The Magic Christmas Tree is a 1964 American\\
    Christmas-themed fantasy-adventure film about\\
    a boy who uses a magic ring to bring a\\
    Christmas tree to life.
} &
\makecell[l]{%
$(\textsf{The Magic Christmas Tree},\textsf{P1071},\textsf{U.S.})$\\
$(\textsf{The Magic Christmas Tree},\textsf{P495},\textsf{U.S.})$
} \\
\addlinespace[5pt]
\multicolumn{4}{l}{\makecell[l]{%
\textbf{Explanation.} The predicate \textsf{location of creation (P1071)} is used to specify the physical location of production of an item, while\\
\textsf{country of origin (P495)} specifies the attributed nationality. While the country of origin is specified by the source text\\
(i.e., by the word ``American'') the location of creation is not, yet triples with both predicates are extracted by the LLM.
}} \\
\midrule
\makecell[l]{%
    \textsf{commercialization date (P5204)}\\
    \textsf{publication date (P577)}\\
} & 242 &
\makecell[l]{%
    Beautiful Eyes is the second extended play (EP)\\
    by American singer-songwriter Taylor Swift.\\
    The EP was released on July 15, 2008 by\\
    Big Machine Records exclusively to Walmart\\
    stores in the United States and online.
} &
\makecell[l]{%
$(\textsf{Beautiful Eyes},\textsf{P5204},\textsf{15 July 2008})$\\
$(\textsf{Beautiful Eyes},\textsf{P577},\textsf{15 July 2008})$
} \\
\addlinespace[5pt]
\multicolumn{4}{l}{\makecell[l]{%
\textbf{Explanation.} The predicate \textsf{commercialization date (P5204)} is typically used to specify the date on which a certain product\\
or invention is firstly commercialized. Instead, \textsf{publication date (P577)} specifies the date on which a creative work such as\\
a musical album is made public. While \textsf{P5204} and \textsf{P577} are substantially different predicates and only one of them\\
should be selected, the LLM extracts two triples with both predicates yet encoding the same fact.
}} \\
\midrule
\makecell[l]{%
    \textsf{location of creation (P1071)}\\
    \textsf{country of origin (P495)}\\
    \textsf{location of formation (P740)}\\
} & 21 &
\makecell[l]{%
    Liberty Tax Service is an American company\\
    specializing in the preparation of tax returns\\
    for individuals and small businesses. [\ldots]\\
    The company began in Canada in 1997 when\\
    John Hewitt, co-founder of Jackson Hewitt,\\
    acquired a Canadian tax franchisor,\\
    U\&R Tax Depot.
} &
\makecell[l]{%
$(\textsf{Liberty Tax, Inc.},\textsf{P1071},\textsf{Canada})$\\
$(\textsf{Liberty Tax, Inc.},\textsf{P495},\textsf{Canada})$\\
$(\textsf{Liberty Tax, Inc.},\textsf{P740},\textsf{Canada})$
} \\
\addlinespace[5pt]
\multicolumn{4}{l}{\makecell[l]{%
\textbf{Explanation.} The predicate \textsf{location of creation (P1071)} is used to specify the physical location of production of an item, while\\
\textsf{country of origin (P495)} specifies the attributed nationality. Instead, \textsf{location of formation (P740)} specifies where an organization was\\
originally established. Instead of selecting only a single predicate expressing the fact that \textsf{Liberty Tax, Inc.} started in \textsf{Canada} (i.e., \textsf{P740}),\\
the LLM selected all predicates and extracted multiple triples encoding the same fact.
}} \\
\bottomrule
\end{tabular}
\caption{\textbf{The \OurInContext baseline with \textsc{Qwen3-30B-A3B} erroneously extracts multiple triples with different predicates to encode a single fact, hence artificially inflating performances on SPARQL patterns} (see \cref{sec:emprical-evaluations-patterns}). We show three sets of predicates that are often used together in multiple triples encoding the same information from the source text. The redundancy of the extracted triples increases the average edge multiplicity, and therefore it artificially inflates the frequencies of basic graph patterns (BGPs) (as shown in \cref{tab:empirical-evaluation-patterns}).}
\label{tab:qualitative-redundant-triples}
\end{table*}

\clearpage%
\section{Prompts}\label{app:prompts}

\subsection{\OurMethod}\label{app:prompts-our-method}

\subsubsection*{Open-domain KG extraction}

\begin{promptbox}
\begin{minted}[breaklines,breaksymbolleft=,fontsize=\small,bgcolor=promptbg]{text}
## Task Description
Your task is to extract a Wikidata-like knowledge graph from text. A knowledge graph consists of subject-predicate-object triples where:
- subject and object: Named entities or concepts that describe groups of people, events, or any abstract objects.
- predicate: A predicate (or relation type) that connects the subject and object.

Additionally, some triples may have qualifiers providing contextual information about a triple (e.g., date, place, or other attributes).
Each qualifier only exists when linked to a triple and consists of a predicate-object pair.

## Inputs
You will receive the text.

## Output Format
Extract triples and qualifiers in **JSON** as a list of dictionaries, where each dictionary contains:
- "triple": A list of three elements: subject, predicate, and object
- "subject_type": The type of the subject
- "object_type": The type of the object
- "qualifiers": An optional list of dictionaries, where each dictionary contains:
    - "pair": A list of two elements: qualifier predicate, and qualifier object
    - "object_type": The type of the qualifier object

## Example
<... an example here ...>
\end{minted}
\end{promptbox}

\subsubsection*{Entity Type Canonicalization}

\begin{promptbox}
\begin{minted}[breaklines,breaksymbolleft=,fontsize=\small,bgcolor=promptbg]{text}
## Task Description
Your task is to disambiguate an entity type for one or more entities extracted from text.

## Inputs
You will receive a text, a list of extracted entities, an extracted entity type, and a list of candidate types.
Return a single type for all the entities. No additional text.
\end{minted}
\end{promptbox}

\newpage%
\subsubsection*{Predicate Canonicalization}

\begin{promptbox}
\begin{minted}[breaklines,breaksymbolleft=,fontsize=\small,bgcolor=promptbg]{text}
## Task Description
Your task is to disambiguate a predicate indicating the relationship type between pairs of entities extracted from text. Assume the predicate direction is not important.

## Inputs
You will receive a text, a list of extracted entity pairs, the extracted predicate, and a list of candidate predicates.
Return a single predicate for all the entity pairs. No additional text.
\end{minted}
\end{promptbox}

\subsubsection*{Entity Deduplication}

\begin{promptbox}
\begin{minted}[breaklines,breaksymbolleft=,fontsize=\small,bgcolor=promptbg]{text}
## Task Description
Find duplicates for a given entity, and an entity alias that best represents the duplicates.
Duplicates are those that are the same in meaning, such as with variation in tense, plural form, stem form, case, abbreviation, shorthand.

## Inputs
You will receive an entity and a list of candidate duplicate entities.

## Output Format
Return duplicates and the alias in **JSON** as a dictionary containing:
- "duplicates": A list of duplicate entities taken from the candidates list
- "alias": Best entity name to represent the duplicates

If there are no duplicates, then return the empty dictionary {}.
\end{minted}
\end{promptbox}

\subsection{\OurMethodCorrect}\label{app:prompts-our-method-corrections}

\subsubsection*{Triple Correction}

\begin{promptbox}
\begin{minted}[breaklines,breaksymbolleft=,fontsize=\small,bgcolor=promptbg]{text}
## Task Description
You are a knowledge graph expert. Your task is to correct an invalid subject-predicate-object (SPO) triple so that it satisfies ontology constraints, while still reflecting information present in the source text.

\end{minted}
\end{promptbox}
\newpage%
\begin{promptbox}
\begin{minted}[breaklines,breaksymbolleft=,fontsize=\small,bgcolor=promptbg]{text}
## Inputs
- Source text
- The invalid triple (subject, predicate, object)
- The domain and range constraints of the predicate
- A list of candidate replacement predicates
- The reasons why the triple is invalid

## Allowed Actions (applied in sequence)
- "swap": swap subject and object
- "add_subject_type": add type to subject (to satisfy domain)
- "add_object_type": add type to object (to satisfy range)
- "replace_predicate": replace predicate with a candidate

## Rules
- Added types must be plausible given the text
- Only use candidates for replace_predicate
- Actions are applied in order, each modifying the result of the previous one
- Choose the sequence of actions that produces a triple consistent with the ontology constraints and grounded in the source text. If no such sequence exists, return an empty list []

## Output
Return only a **JSON** list of [action, value] pairs, where action is one of swap, add_subject_type, add_object_type, replace_predicate, and value is the new type or predicate string (or null for swap). Return an empty list if no correction is possible.

### Output Examples
- [["replace_predicate", "publication date"]]
- [["swap", null], ["add_object_type", "person"]]
- []
\end{minted}
\end{promptbox}

\subsubsection*{Qualifier Correction}

\begin{promptbox}
\begin{minted}[breaklines,breaksymbolleft=,fontsize=\small,bgcolor=promptbg]{text}
## Task Description
You are a knowledge graph expert. Your task is to correct a qualifier predicate-object pair associated to a triple so that it satisfies ontology constraints, while still reflecting information present in the source text.

\end{minted}
\end{promptbox}
\newpage%
\begin{promptbox}
\begin{minted}[breaklines,breaksymbolleft=,fontsize=\small,bgcolor=promptbg]{text}
## Inputs
- Source text
- The triple (subject, predicate, object)
- The invalid qualifier (qualifier predicate, qualifier object)
- The range constraint of the qualifier predicate
- A list of candidate replacement qualifier predicates
- The reason why the qualifier is invalid

## Allowed Actions
- "add_object_type": add type to qualifier object (to satisfy range)
- "replace_predicate": replace qualifier predicate with a candidate

## Rules
- Added types must be plausible given the text
- Only use candidates for replace_predicate
- Choose the action that produces a qualifier consistent with the ontology constraints and grounded in the source text. If no action exists, return an empty list []

## Output
Return only a **JSON** list [action, value], where action is one of add_object_type, replace_predicate, and value is the new type or predicate string. Return an empty list if no correction is possible.

### Output Examples
- ["replace_predicate", "doctoral advisor"]
- ["add_object_type", "person"]
- []
\end{minted}
\end{promptbox}

\subsection{\OurInContext Baseline}\label{app:prompts-in-context-baseline}

\subsubsection*{Extract Predicates}

\begin{promptbox}
\begin{minted}[breaklines,breaksymbolleft=,fontsize=\small,bgcolor=promptbg]{text}
## Task Description
Extract Wikidata-like predicates (or relationship types) explaining the relationships between the entities in the given text.

## Inputs
You will receive the text.

## Output Format
Return the predicates in **JSON** as a list of strings.

\end{minted}
\end{promptbox}
\newpage%
\begin{promptbox}
\begin{minted}[breaklines,breaksymbolleft=,fontsize=\small,bgcolor=promptbg]{text}
## Example
Text:
Alan Turing (23 June 1912 - 7 June 1954) was an English mathematician. He received the Smith's Prize in 1936, and in 1938 he earned a doctorate degree from Princeton University, advised by Alonzo Church.

### Output
[
    "date of birth",
    "date of death",
    "country of citizenship",
    "occupation",
    "award received",
    "educated at"
]
\end{minted}
\end{promptbox}

\subsubsection*{Extract the Knowledge Graph}

\begin{promptbox}
\begin{minted}[breaklines,breaksymbolleft=,fontsize=\small,bgcolor=promptbg]{text}
## Task Description
Your task is to extract a Wikidata-like knowledge graph from text. A knowledge graph consists of subject-predicate-object triples where:
- subject and object: Named entities or concepts that describe groups of people, events, or any abstract objects.
- predicate: A predicate (or relation type) that connects the subject and object.

Additionally, some triples may have qualifiers providing contextual information about a triple (e.g., date, place, or other attributes).
Each qualifier only exists when linked to a triple and consists of a predicate-object pair.

### Task Constraints
You are allowed to use only the provided predicates.
Most importantly, each provided predicate comes with its own set of allowed domain and range entity types.
For each triple, the subject type must be one (or a specialization) of any type in the domain list specified by the predicate.
In addition, the object type must be one (or a specialization) of any type in the range list specified by the predicate.
If no types are specified as domain or range, then there are no type constraints for the subject or object entity, respectively.

\end{minted}
\end{promptbox}
\newpage%
\begin{promptbox}
\begin{minted}[breaklines,breaksymbolleft=,fontsize=\small,bgcolor=promptbg]{text}
Furthermore, you are allowed to use only the provided qualifiers.
Most importantly, each predicate comes with its own list of allowed qualifier predicates.
In addition, the qualifier object type must be one (or a specialization) of any type in the range list specified by the qualifier predicate.
If no types are specified as range, then there are no type constraints for the qualifier object entity.

## Inputs
You will receive the text, the list of allowed predicates each with their own domain-range and permitted qualifiers constraints, and the list of allowed qualifiers each with their own range constraint.

## Output Format
Extract triples and qualifiers in **JSON** as a list of dictionaries, where each dictionary contains:
- "triple": A list of three elements: subject, predicate, and object
- "subject_type": The type of the subject
- "object_type": The type of the object
- "qualifiers": An optional list of dictionaries, where each dictionary contains:
    - "pair": A list of two elements: qualifier predicate, and qualifier object
    - "object_type": The type of the qualifier object

## Example
<... an example here ...>
\end{minted}
\end{promptbox}

\end{document}